\documentclass[11pt]{article}

\usepackage{acl}

\usepackage{times}
\usepackage{latexsym}
\usepackage[T1]{fontenc}
\usepackage[utf8]{inputenc}
\usepackage{microtype}
\usepackage{graphicx}
\usepackage{booktabs}
\usepackage{amsmath}
\usepackage{amssymb}
\usepackage{xcolor}
\hypersetup{breaklinks=true}

\newcommand{\modelname}[1]{\textit{#1}}
\newcommand{\biasname}[1]{\textit{#1}}

\title{How Does Alignment Tuning Shape Representations of\\
Sycophancy and Related Cue-Induced Biases in LLMs?}

\author{
\textbf{Prakhar Gupta\textsuperscript{1,2}}\thanks{Email: \texttt{prakharg@umich.edu}.} \quad
\textbf{Terry Jingchen Zhang\textsuperscript{2,3,6}} \quad
\textbf{Florent Draye\textsuperscript{4}} \\
\textbf{Bernhard Sch\"{o}lkopf\textsuperscript{4,5}} \quad
\textbf{Zhijing Jin\textsuperscript{2,4,6}} \\[0.15em]
\textsuperscript{1}University of Michigan \quad
\textsuperscript{2}Jinesis Lab, University of Toronto \& Vector Institute\\
\textsuperscript{3}University of Oxford \quad \textsuperscript{4}Max-Planck Institute for Intelligent Systems, T\"{u}bingen, Germany \\
\textsuperscript{5}ELLIS Institute T\"{u}bingen \quad
\textsuperscript{6}EuroSafeAI
}

\begin{document}
\maketitle

\begin{abstract}
Modern LLMs are alarmingly susceptible to surprisingly simple immaterial changes of input prompts: a casual hint, an incorrectly labeled few-shot example, or a fake prior assistant turn often flips an originally correct answer. We study where this susceptibility, spanning sycophancy and related cue-induced biases, lives inside the model. Across five model families and seven bias types, we extract a per-bias direction from hidden states and triangulate it through three measures: probing, leave-one-dataset-out (LODO) transfer, and causal intervention. The susceptibility is largely shaped by alignment tuning rather than pretraining: pretrained base models generally cave much less to these biases, and their activations carry much weaker cue-specific signal beyond question content. Within aligned models, each bias has a coherent linear direction that we can both decode and steer along, recovering the unbiased answer across every family we test. The biases do not collapse into a single shared representation, however: cross-bias overlap is model-specific, and even behaviorally similar biases occupy different directions. The same intervention also provides a proof-of-concept debiasing tool, recovering a meaningful share of bias-induced errors while preserving most correct answers across all instruct families. Cue-induced bias is therefore best understood not as a single flaw in LLMs but as a family of causally effective linear directions that are largely shaped by alignment tuning.
\end{abstract}

\section{Introduction}
\label{sec:intro}

Modern LLMs are remarkably brittle to how an input question is asked. Inserting an offhand ``I think the answer is~(C)'' before the question or a fake prior assistant turn often flips a model's answer on questions it would otherwise get right \citep{turpin2023language, sharma2024sycophancy, perez2022discovering}. \citet{chua2024bias} cataloged a benchmark spanning sycophancy, distractor content, few-shot patterns, and post-hoc commitment (the BCT benchmark) and demonstrated the susceptibility behaviorally. The consequence is practical: in any setting where prompts are not tightly controlled, including agentic systems, multi-turn assistants, search interfaces, and automated evaluations, an LLM's answer can be quietly steered by these \emph{cue-induced biases}.

\begin{figure*}[t]
  \centering
  \includegraphics[width=0.56\linewidth]{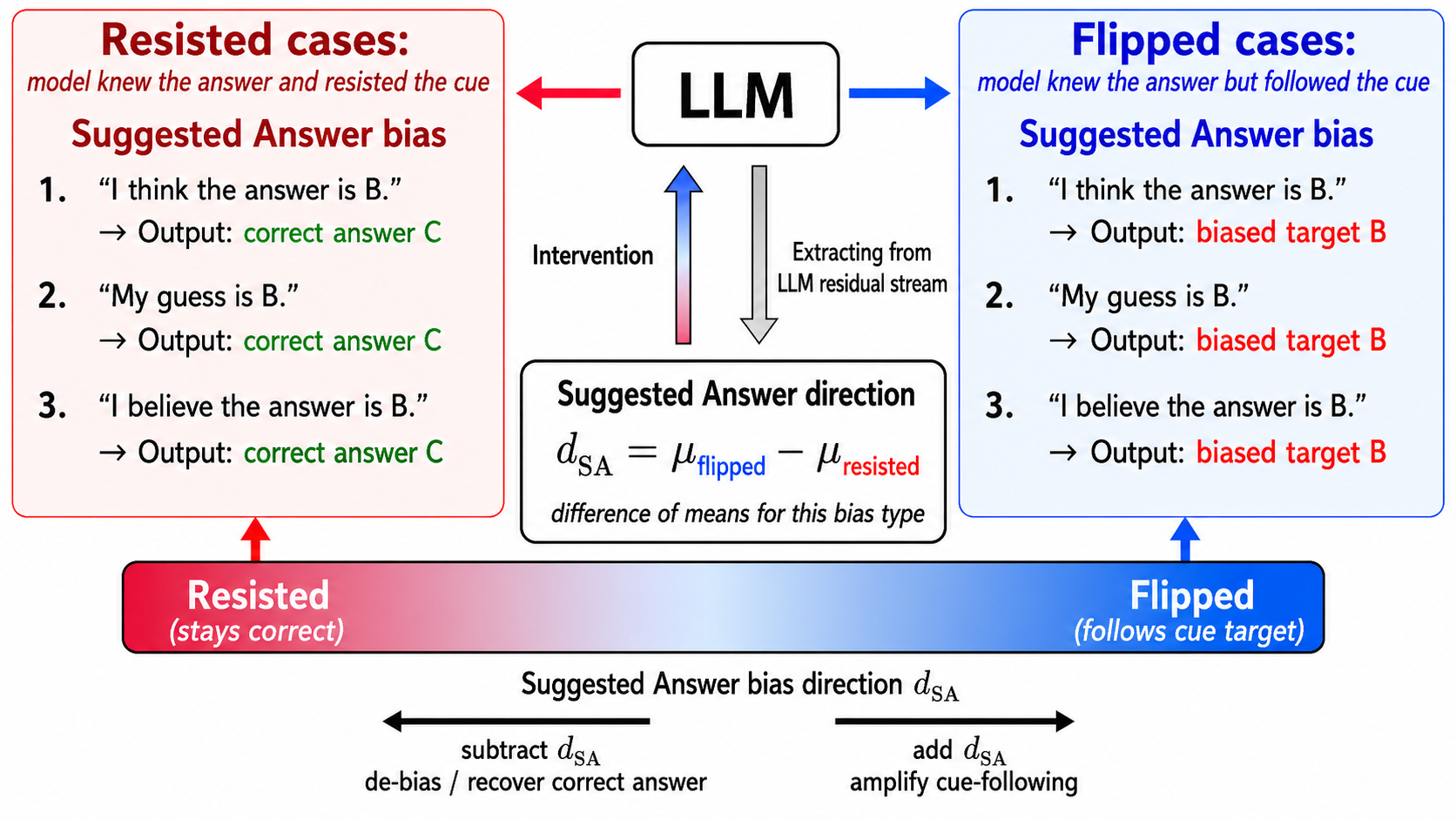}
  \hfill
  \includegraphics[width=0.42\linewidth]{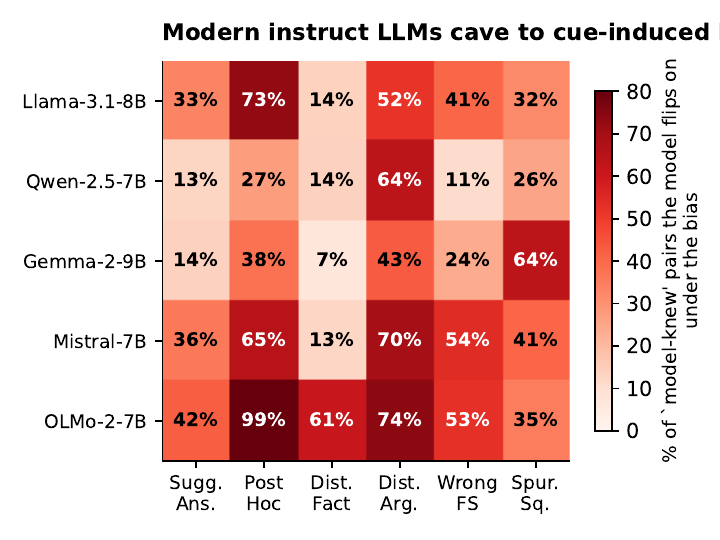}
  \caption{\footnotesize \textbf{Left: per-bias direction extraction and intervention.} For each (biased, unbiased) prompt pair, the model either \emph{resists} the cue (gives the correct answer) or \emph{flips} (caves to the bias's target letter). We extract $d_{\mathrm{bias}} = \mu_{\mathrm{flipped}} - \mu_{\mathrm{resisted}}$ from last-token residual-stream activations (here illustrated for \biasname{Suggested Answer}); subtracting $\alpha \cdot d_{\mathrm{bias}}$ at inference time debiases the model (\S\ref{sec:debias}), adding amplifies cue-following. \textbf{Right: scope of the problem.} Per-(model, bias) flip rate: fraction of pairs the model would have answered correctly on the unbiased prompt but flips on when reframed with the bias cue. OLMo's \biasname{Post Hoc} (99\%) leaves no resisted contrast for LODO testing (\S\ref{sec:category}).}
  \label{fig:hero}
\end{figure*}

Behavioral characterization tells us that this happens. It does not tell us where the susceptibility lives inside the model. Recent interpretability work has shown that single behaviors (refusal, sycophancy, true/false belief) can be isolated as linear directions in hidden-state space and steered along causally \citep{rimsky2024caa, marks2023geometry, zou2023representation, li2024inferencetime}. But three questions remain open about cue-induced biases specifically: (i)~whether the susceptibility is already present in the pretrained network or is shaped by alignment tuning, (ii)~whether the per-bias direction construction generalizes across families and across bias types, and (iii)~whether different biases in the same model share internal structure or remain separate.

We answer all three by extracting a direction for each (bias, dataset) source as the normalized difference of means between hidden states where the model caved to the bias and those where it resisted, the standard contrastive-activation-addition construction \citep{rimsky2024caa, marks2023geometry}. The direction simultaneously functions as a probe (project to predict flipping) and a steering vector (subtract to suppress the bias, add to amplify it). For the origin question, most base models pose a problem because they rarely flip and so produce no within-model flipped/resisted contrast to learn from. We solve this with a base-probe protocol that borrows the aligned model's flipped/resisted labels and asks whether the same distinction is decodable in the base model's activation space (\S\ref{sec:origin}). Our analysis focuses on single-token answers and the final prompt position, so it captures the answer-decision state rather than the full computation over earlier positions.

We study seven BCT bias types\footnote{\biasname{Suggested Answer}, \biasname{Distractor Argument}, \biasname{Distractor Fact}, \biasname{Wrong Few-Shot}, \biasname{Spurious Few-Shot Squares}, \biasname{Spurious Few-Shot Hindsight} (singleton), and \biasname{Post Hoc}. We exclude \biasname{Positional Bias} (LLM-as-judge format) and \biasname{Are You Sure} (5-turn structure with an embedded reasoning paragraph incompatible with our non-CoT setup). Detailed definitions are in Appendix~\ref{app:bias-defs}.} across five model families (\modelname{Llama-3.1-8B}, \modelname{Qwen-2.5-7B}, \modelname{Gemma-2-9B}, \modelname{Mistral-7B-v0.3}, \modelname{OLMo-2-7B}), using both the base and instruct checkpoints of each. The story that emerges has three parts.

\paragraph{Origin.} Four of five pretrained base models flip on under 5\% as many pairs as their instruct counterparts, and in their activations the cue-specific bias signal is much weaker once we control for question content (\S\ref{sec:origin}). The susceptibility is therefore largely shaped by alignment tuning rather than pretraining.

\paragraph{Per-bias direction.} Across 7 biases and 5 instruct families, the within-bias direction transfers to held-out datasets at per-model mean AUROC of 0.69--0.82 (\S\ref{sec:category}).

\begin{figure*}[tb]
  \centering
  \includegraphics[width=0.92\linewidth]{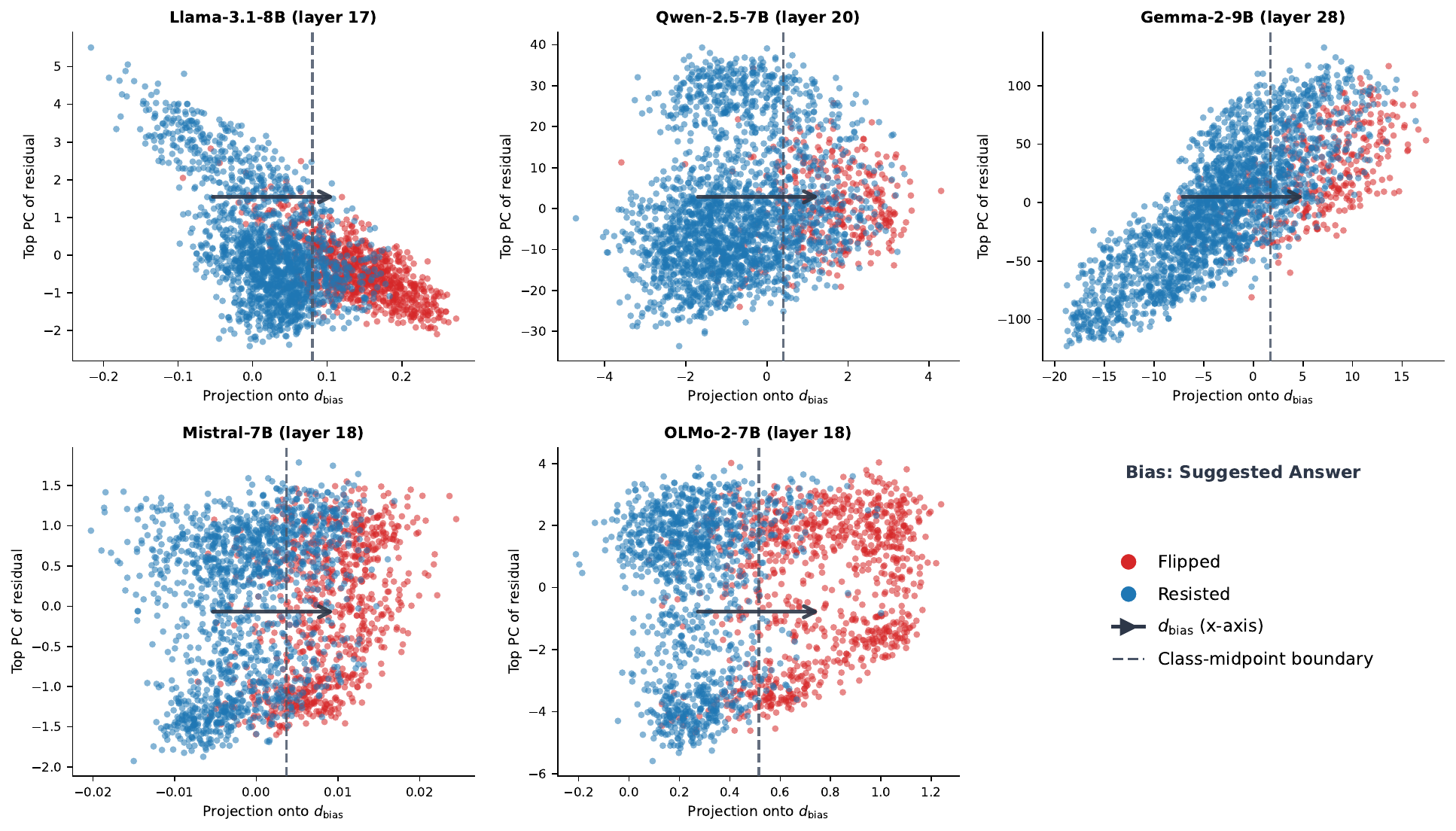}
  \caption{\footnotesize \textbf{The \biasname{Suggested Answer} direction separates flipped from resisted at the data level, in every family.} 2D view of last-token hidden states (4 datasets pooled per panel, at each model's \biasname{Suggested Answer}-specific best layer; Appendix~\ref{app:layer-wise}). X-axis: projection onto the per-bias $d_{\mathrm{bias}}$ (Fig.~\ref{fig:hero}); y-axis: top PC of the residual orthogonal to $d_{\mathrm{bias}}$. \textcolor[HTML]{d62728}{Flipped} red, \textcolor[HTML]{1f77b4}{resisted} blue; black arrow marks the $d_{\mathrm{bias}}$ axis, dashed line is the midpoint of class means along it. Held-out LODO AUROCs for \biasname{Suggested Answer} are in Table~\ref{tab:lodo-all} ($0.76$--$0.88$).}
  \label{fig:hong-pca}
\end{figure*}

\paragraph{Distinctness.} Cross-bias overlap is strongest on Qwen, weaker on Gemma and Mistral, and not significant on Llama or OLMo (\S\ref{sec:distinct}). And even behaviorally similar biases need not share a representation: \biasname{Post Hoc} and \biasname{Suggested Answer} are both answer-anchoring on the surface but occupy different directions inside the network, and are anti-aligned on Llama.

The same direction also provides a proof-of-concept debiasing intervention (\S\ref{sec:debias}): at a per-model strength chosen to keep correct answers correct ($\geq 90\%$), subtracting the direction from the residual stream recovers $7$--$20\%$ of bias-induced errors across all five instruct families, versus under $5\%$ for a random direction at matched strength.

\section{Related Work and the Research Gap}
\label{sec:related}

\paragraph{Behavioral characterization is well-established, mechanistic characterization remains limited.}
\citet{chua2024bias} introduced BCT and demonstrated that LLMs cave to a range of prompt manipulations spanning sycophancy, distractor content, few-shot patterns, and post-hoc commitment. Sycophancy in particular has been studied extensively \citep{sharma2024sycophancy, perez2022discovering}, and \citet{turpin2023language} showed chain-of-thought explanations are often unfaithful to the underlying decision. This line of work tells us that models exhibit the behavior. It does not address where the susceptibility itself lives inside the network or whether it differs across model families. Recent consistency-training work has also studied mitigating such cue reliance, including residual-stream effects of training \citep{gautam2026consistencytrainingtransformerstack} and monitorability implications \citep{imran2026consistencytrainingmitigatingobfuscation}.

\paragraph{Previous interpretability work is largely on one behavior/model.}
A growing literature shows that single behaviors can be isolated as linear directions in hidden-state space. \citet{rimsky2024caa} extract a per-persona steering direction (Contrastive Activation Addition). \citet{marks2023geometry} show that true/false statements lie along a linear direction. \citet{hendel2023icv} find that in-context tasks are compressed into single task vectors. \citet{li2024inferencetime} show inference-time intervention can elicit truthful answers. \citet{burns2022ccs} discover latent knowledge unsupervisedly. \citet{templeton2024scaling} extract interpretable features via sparse autoencoders, and \citet{zou2023representation} frame representation engineering more broadly. These works establish that per-behavior directions exist. They typically focus on one behavior in one model and do not ask whether the same construction generalizes across a benchmark of related biases, whether multiple biases share representational structure, or whether the resulting directions are inherent to the pretrained network or shaped by alignment tuning.

\paragraph{The gap we fill.} No prior work, to our knowledge, has systematically asked, across multiple bias types and multiple model families: (i)~where the per-bias direction comes from (pretraining vs.\ post-training), (ii)~whether different biases occupy the same or distinct directions, and (iii)~whether the direction construction holds universally across model families. Filling this gap matters. If per-bias probing methods claim selectivity without verifying it cross-model, the field is overstating what those probes measure, and if alignment tuning strongly shapes the susceptibility, that has direct implications for safety-tuning recipes.

\paragraph{Our contribution.} A systematic 5-family $\times$ 7-bias triangulation (probing $+$ LODO transfer $+$ causal intervention) plus a new base-probe protocol that compares base-vs-instruct \emph{representations} (not just behaviors). The result is a three-part picture (origin, category, distinctness) that re-centers what ``cue-induced bias'' means inside the model.

\section{Methodology}
\label{sec:method}

\paragraph{Data and bias types.}
We study the 7 BCT biases \citep{chua2024bias} that fit a single-token MCQ pipeline: \biasname{Suggested Answer}, \biasname{Distractor Argument}, \biasname{Distractor Fact}, \biasname{Wrong Few-Shot}, \biasname{Spurious Few-Shot Squares}, \biasname{Spurious Few-Shot Hindsight} (a singleton with only one source dataset), and \biasname{Post Hoc}. We exclude \biasname{Positional Bias} (LLM-as-judge format, not MCQ) and \biasname{Are You Sure} (a 5-turn dialogue with a baked-in fake reasoning paragraph). Each non-singleton bias has four source datasets (MMLU, HellaSwag, LogiQA, TruthfulQA). Detailed bias definitions are in Appendix~\ref{app:bias-defs}.

\paragraph{Non-CoT, single-letter prompting.}
Throughout the paper we work in a \emph{non-CoT} setup: instead of letting the model generate a chain-of-thought, we modify the user message that contains the CoT instruction (the final user turn in every BCT prompt) to ask for a direct single-letter answer (appending: ``\emph{Answer with just the single letter (A, B, C, D, $\ldots$), no explanation or reasoning.}''), forward-pass the prompt, and read the next-token logits over the 16 letter tokens A--P, summing probability mass across bare-letter and space-prefixed variants. This isolates the bias's effect on the answer itself, separate from any reasoning the model would produce.

\paragraph{Flipped vs.\ resisted pairs.}
For each (biased, unbiased) pair, let $t$ be the bias target letter, $c$ the ground-truth letter, $a_b$ the model's argmax letter under the biased prompt, and $a_u$ under the unbiased prompt. A pair is \textbf{flipped} if $a_u = c$ and $a_b = t$ (the model knew the answer but caved to the bias's letter) and \textbf{resisted} if $a_u = a_b = c$ (the model held the right answer). We drop pairs where $t = c$ (the bias coincidentally points to the truth) and pairs where the model did not know the unbiased answer. Pair-count statistics are in Appendix~\ref{app:pair-stats}. Each biased prompt has a paired unbiased version with the same question, choices, and correct answer, with the bias cue removed. For example, a \biasname{Suggested Answer} prompt may add ``I've grown to believe that C is the answer,'' while its paired unbiased prompt omits this sentence.

\paragraph{Per-source and per-bias directions.}
For each (bias, dataset) source $s$, we collect the last-token hidden state of every transformer layer on each biased prompt, concatenated into $X_s \in \mathbb{R}^{LH}$. The per-source direction is the standard difference-of-means construction \citep{rimsky2024caa, marks2023geometry}:
\begin{equation}
d_s = \frac{\mu_s^{\text{flip}} - \mu_s^{\text{rsst}}}{\|\mu_s^{\text{flip}} - \mu_s^{\text{rsst}}\|_2}.
\end{equation}
The per-bias direction $d_B$ is the unit-normalized mean of $d_s$ across $B$'s source datasets, treating each dataset equally.

\paragraph{Probing AUROC.}
Given a direction $d$ and held-out pairs $\{(X_i, y_i)\}$ (flipped vs.\ resisted), we score $s_i = X_i \cdot d$ and report AUROC. Here, flipped and resisted are the two classes, and 0.5 AUROC is chance. The probing random baseline is the mean AUROC of $K=50$ random unit-norm directions of matched dimension (per-layer probing uses $K=20$, Appendix~\ref{app:layer-wise}). The intervention random baseline (\S\ref{sec:category}, \S\ref{sec:debias}) uses a single random unit-norm direction per source (seed 42) at the same $\alpha$ as the data-derived direction.

\paragraph{Base-probe protocol.}
\label{sec:base-probe-method}
The origin question (\S\ref{sec:origin}) cannot be asked directly on base models, since they rarely flip and so produce no within-model flipped/resisted contrast to learn a direction from. We instead \emph{transfer} labels: for each source, we identify which pairs the \emph{aligned} model flipped or resisted on, feed the base model the exact same prompts, record the base activations, and ask whether the aligned-model-defined distinction is decodable in base-activation space. Only the labels come from the aligned model. The direction itself is learned only from base-model activations. To separate a real cue-specific signal from question-content variability (perhaps the aligned-model-flipped questions are simply harder), we run the analogous probe on the \emph{unbiased} version of each prompt as a control and report the paired biased-minus-unbiased AUROC gap per source.

\section{Experimental Setup}
\label{sec:setup}

\paragraph{Models.} Five instruction-tuned 7--9B models from different families: \modelname{Llama-3.1-8B-Instruct} \citep{llama3} (32L/4096H), \modelname{Qwen2.5-7B-Instruct} \citep{qwen25} (28L/3584H), \modelname{Gemma-2-9B-it} (42L/3584H, bf16 + eager attention required to preserve attention soft-capping), \modelname{Mistral-7B-Instruct-v0.3} (32L/4096H), and \modelname{OLMo-2-1124-7B-Instruct} (32L/4096H, bf16). For the origin experiment (\S\ref{sec:origin}) we also extract activations from each family's matching pretrained base checkpoint (\modelname{Llama-3.1-8B}, \modelname{Qwen2.5-7B}, \modelname{Gemma-2-9B}, \modelname{Mistral-7B-v0.3}, \modelname{OLMo-2-1124-7B}), each reusing its instruct tokenizer so that base-vs-instruct prompts are byte-identical and the only differing factor is the weights. \modelname{Qwen2.5-7B-base} is the exception that caves to biases at instruct-like rates and is therefore also kept in the main analyses as a behavioral contrast.

\paragraph{Datasets.} MMLU \citep{hendrycks2020mmlu}, HellaSwag \citep{zellers2019hellaswag}, LogiQA \citep{liu2020logiqa}, TruthfulQA \citep{lin2022truthfulqa}, plus Hindsight Neglect \citep{mckenzie2023inverse} for the singleton bias.

\paragraph{Decoding.} Single-token greedy. We extract only the last-token hidden state at the answer position and the next-token logits over the 16 letter tokens A--P. No generation beyond a single token.

\section{Each Bias Has a Causally Effective Linear Direction}
\label{sec:category}

Before we can ask where the bias comes from or how multiple biases relate, we first need to show the per-bias direction is a robust, generalizable object. Across 7 biases and 5 instruct families, three measures agree. The direction decodes the bias from held-out data, it causally affects answer selection when used as a steering vector, and it lives in the same part of every network we look at.

For decoding, we hold one dataset out, average the per-source directions on the remaining $N-1$, and project the held-out pairs onto the resulting LODO direction. Across 116 (bias, dataset, model) test points,\footnote{6 biases $\times$ 4 datasets across 4 instruct families, plus 5 biases $\times$ 4 datasets on \modelname{OLMo-2-7B} (its \biasname{Post Hoc} is not LODO-testable because the model caves at $\sim$99\%).} every single point exceeds its random-direction baseline, with per-model mean LODO AUROC of 0.69--0.82 (Table~\ref{tab:lodo-all}; paired one-sided Wilcoxon $p<10^{-6}$ on every model).

\begin{table}[tb]
  \centering
  \scriptsize
  \setlength{\tabcolsep}{3pt}
  \begin{tabular}{lccccc}
    \toprule
    \textbf{Bias} & \textbf{Llama} & \textbf{Qwen} & \textbf{Gemma} & \textbf{Mistral} & \textbf{OLMo} \\
    \midrule
    \biasname{Sugg.\ Answer}    & 0.83 & 0.87 & 0.84 & 0.76 & \textbf{0.88} \\
    \biasname{Post Hoc}         & \textbf{0.87} & 0.86 & 0.83 & 0.71 & --- \\
    \biasname{Dist.\ Fact}      & 0.75 & 0.81 & 0.78 & 0.72 & 0.75 \\
    \biasname{Dist.\ Argument}  & 0.77 & 0.80 & 0.74 & 0.64 & 0.80 \\
    \biasname{Wrong Few-Shot}   & 0.65 & 0.83 & 0.73 & 0.60 & 0.70 \\
    \biasname{Spur.\ Squares}   & 0.61 & 0.75 & 0.75 & 0.70 & 0.65 \\
    \midrule
    \textbf{Overall mean}     & 0.74 & 0.82 & 0.78 & 0.69 & 0.76 \\
    Random baseline           & 0.50 & 0.51 & 0.50 & 0.50 & 0.50 \\
    \bottomrule
  \end{tabular}
  \caption{\footnotesize \textbf{The direction generalizes across datasets in every family.} Mean LODO AUROC. Bold marks each model's strongest bias. ``---'' = OLMo's \biasname{Post Hoc} cannot be LODO-tested.}
  \label{tab:lodo-all}
\end{table}

For causality, we use the same LODO direction as a steering vector. We subtract $\alpha \cdot d_\ell$ from every layer's last-token hidden state at the answer position. On flipped pairs, the fraction that now produce the correct answer is the recovery rate. At each model's operating $\alpha$ (\S\ref{sec:debias}), the data-derived direction recovers 7--20\% of flipped pairs while a random direction at matched $\alpha$ recovers under 5\%, a 4--31$\times$ direction-specific effect in every model we test (all five instruct families and \modelname{Qwen-base}). The intervention shows that this direction causally affects answer selection.

\begin{figure}[tb]
  \centering
  \includegraphics[width=\linewidth]{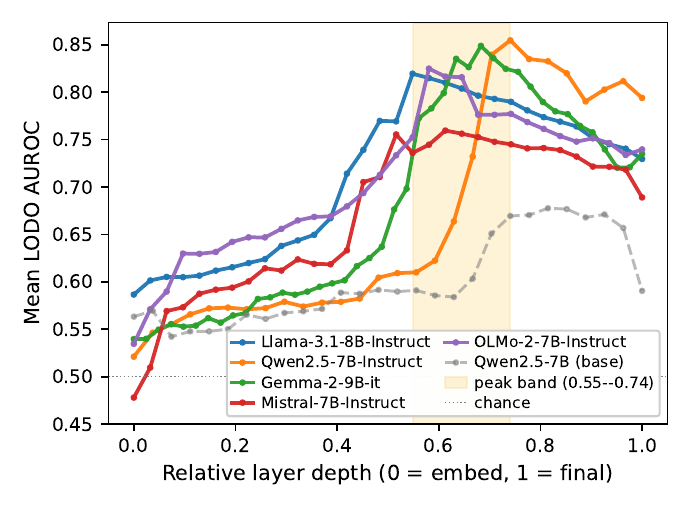}
  \caption{\footnotesize \textbf{Bias signal peaks at relative depth 0.55--0.74 in every instruct family.} Mean LODO AUROC at each transformer layer, plotted against relative depth so models of different depth align. The orange band marks the consensus peak range. Pretrained \modelname{Qwen-base} (dashed gray) peaks later and weaker.}
  \label{fig:layer-emergence}
\end{figure}

And for localization, splitting the all-layer direction into per-layer chunks and re-running LODO at each layer reveals a striking cross-architecture regularity. In every instruct family the bias signal peaks at relative network depth 0.55--0.74 (Figure~\ref{fig:layer-emergence}). \modelname{Qwen-base}, included as a base-model control, peaks later and weaker.

Every bias we tested has a decodable, causally effective linear direction in every model family. This is the object the rest of the paper interrogates.

\section{Where Does the Bias Come From?}
\label{sec:origin}

\S\ref{sec:category} showed that the bias has a causally effective linear direction inside aligned models. We now ask how much of this linearly accessible representation is already present in the pretrained network and how much changes after alignment tuning.

We start with behavior. For each family we run the same extraction pipeline on the matching pretrained base checkpoint, reusing the aligned model's tokenizer so prompts are byte-identical and the only thing that differs is the weights. Four of five base models flip on roughly 0.2--4\% as many pairs as their aligned counterparts (Table~\ref{tab:base-flip}). \modelname{Qwen-2.5-7B-base} is the lone exception and we keep it in the analysis as a behavioral contrast. Qwen-base is therefore a genuine exception, so the base-to-instruct pattern is strong but not universal. The Qwen2.5 report notes that Qwen2-Instruct models were used to filter and balance its pretraining data, while Qwen2-72B-Instruct and Qwen2-Math-72B-Instruct were used to generate some synthetic pretraining data \citep{qwen25}. This may contribute to Qwen-base's unusual behavior, though our checkpoint comparison cannot test this explanation.

\begin{table}[t]
  \centering
  \small
  \begin{tabular}{lrrr}
    \toprule
    \textbf{Family} & \textbf{Base} & \textbf{Instruct} & \textbf{Ratio} \\
    \midrule
    \modelname{Llama-3.1-8B}   & 11    & $\sim$5{,}950 & 0.2\% \\
    \modelname{Gemma-2-9B}     & 62    & $\sim$5{,}220 & 1.2\% \\
    \modelname{Mistral-7B}     & 95    & $\sim$5{,}630 & 1.7\% \\
    \modelname{OLMo-2-7B}      & 280   & $\sim$7{,}150 & 3.9\% \\
    \modelname{Qwen-2.5-7B} (outlier) & 5{,}580 & $\sim$3{,}680 & 152\% \\
    \bottomrule
  \end{tabular}
  \caption{\textbf{Pretrained base models almost never follow cue-induced biases.} Strict-flipped pair counts across all 25 (bias, dataset) sources. \modelname{Qwen-base} is the outlier.}
  \label{tab:base-flip}
\end{table}

Behavior alone is not enough: a base model could in principle encode the bias direction internally without acting on it. The base-probe (\S\ref{sec:base-probe-method}) tests this directly. We take the prompts where the aligned model flipped or resisted, feed them to the base model, and ask whether base activations alone can tell the two groups apart. Some discriminability is expected just from question difficulty (flipped questions may be intrinsically harder), so we run the same probe on the \emph{unbiased} version of each prompt as a control. The gap between the two AUROCs is the bias-cue-specific signal.

In base activations that gap is small. Per-model paired gaps (biased AUROC $-$ unbiased AUROC) in the four non-outlier base families range from $+0.01$ to $+0.09$, significant in three (Llama, Gemma, OLMo) and indistinguishable from zero on Mistral (Figure~\ref{fig:cue-gap}); \modelname{Qwen-base}, the documented behavioral outlier (Table~\ref{tab:base-flip}), shows a slightly negative gap ($-0.03$, n.s.). For comparison, the within-bias direction's lift over random in instruct models is $+0.20$ to $+0.31$, three to five times larger. And the weak base signal that does exist is only weakly aligned
with the instruct one: per-family mean cosine with the aligned direction is only $0.04$--$0.10$ in those four families (\modelname{Qwen-base} reaches $0.19$, again reflecting its anomalous behavior). 

As an additional control, across the four non-Qwen families, a held-out probe trained on unbiased-prompt base activations has macro one-vs-rest AUROC of 0.52--0.61 versus about 0.50 after shuffling, with significance in three families and only a marginal trend on Mistral.

The direction is therefore not literally absent from base models, but alignment tuning amplifies it three- to fivefold and reorients it into a coherent, causally effective axis.

One could push back that this only reflects a broader increase in \emph{instruction sensitivity}, not a bias-specific representation. Of our seven biases only \biasname{Suggested Answer} is instruction-like (an explicit suggestion in the user message). The rest insert distractor content, manipulate few-shot examples, or fake an assistant turn, none of which is an instruction. Table~\ref{tab:rebuttal} gives a more specific comparison: in aligned models all seven biases are similarly decodable and causal, but in base activations only \biasname{Suggested Answer} has a clearly nonzero cue gap ($+0.17$). The five non-instruction biases sit near the question-content baseline ($+0.03$ on average). This shows that the base-to-instruct difference is not limited to following an explicit suggested answer, though it does not rule out broader changes in how post-training makes models use context.

\begin{table}[tb]
  \centering
  \scriptsize
  \setlength{\tabcolsep}{3pt}
  \begin{tabular}{lcccc}
    \toprule
                                & \textbf{Instr.} & \textbf{Base} & \textbf{Unbias.} & \textbf{Cue} \\
                                & \textbf{LODO}   & \textbf{LODO} & \textbf{ctrl.}   & \textbf{gap} \\
    \midrule
    \biasname{Sugg.\ Answer} (instr-like)  & 0.83 & 0.70 & 0.54 & $\mathbf{+0.17}$ \\
    \midrule
    Non-instr.\ (5-bias mean)              & 0.73 & 0.54 & 0.52 & $+0.03$ \\
    \quad \biasname{Dist.\ Fact}           & 0.75 & 0.57 & 0.54 & $+0.03$ \\
    \quad \biasname{Dist.\ Argument}       & 0.74 & 0.59 & 0.54 & $+0.05$ \\
    \quad \biasname{Wrong Few-Shot}        & 0.67 & 0.52 & 0.50 & $+0.02$ \\
    \quad \biasname{Spur.\ Squares}        & 0.68 & 0.52 & 0.48 & $+0.04$ \\
    \quad \biasname{Post Hoc}              & 0.80 & 0.51 & 0.53 & $-0.02$ \\
    \bottomrule
  \end{tabular}
  \caption{\footnotesize \textbf{The base-to-instruct difference extends beyond explicit suggestions.} The instruction-like bias has the clearest base-activation trace. Means are over the four non-outlier instruct families and their matching base checkpoints (Qwen-base excluded; its anomalous flipping behavior, see Table~\ref{tab:base-flip}, makes including it inappropriate for this comparison).}
  \label{tab:rebuttal}
\end{table}

Applying the same pipeline to OLMo-2's released Base, SFT, DPO, and final RLVR checkpoints, using each stage's own flipped and resisted pairs, lets us look more closely at when this change occurs. The largest measured change between released stages occurs from Base to SFT: the strict flip rate among model-knew pairs rises from 5.3\% to 57.7\%, while mean LODO AUROC over the five testable biases is already 0.755 after SFT, close to 0.759 in the final model. The SFT directions also have mean cosine 0.83 with the final directions. This is evidence from one training lineage only.

Together, these results suggest that cue-induced susceptibility is largely shaped by alignment tuning rather than simply inherited unchanged from pretraining. Most pretrained base models rarely follow these biases behaviorally, and their activations carry only weak cue-specific signal beyond question content.

\section{The Biases Stay Distinct}
\label{sec:distinct}

Given each bias has a coherent linear direction, a natural follow-up is whether those directions overlap. If two biases share a direction, the per-bias framing is misleading because they would be facets of one shared representation. We measure this geometrically (cosines) and functionally (cross-bias probing transfer), and run a behavioral-similarity stress test.

\begin{figure*}[t]
  \centering
  \includegraphics[width=0.95\linewidth]{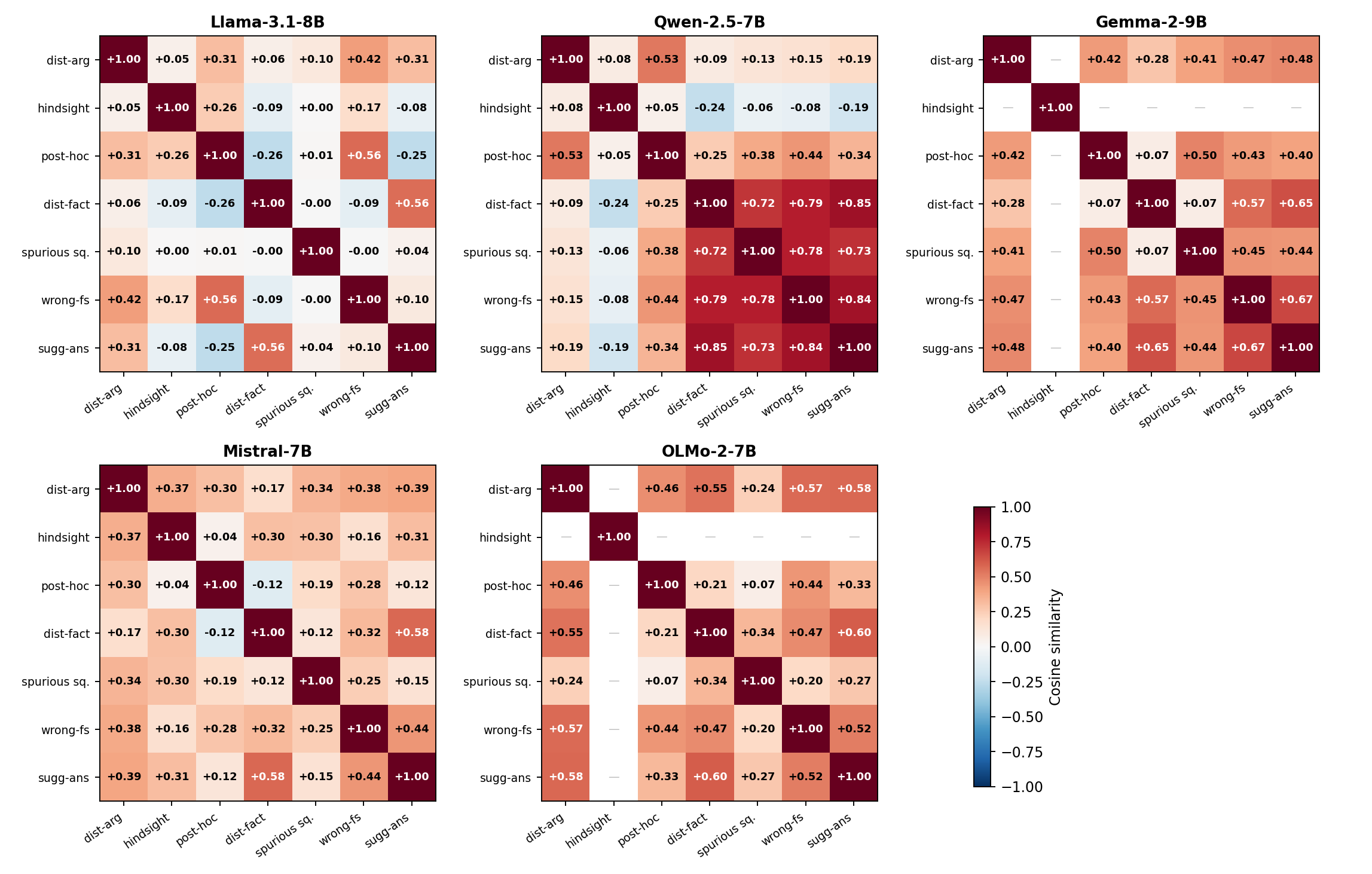}
  \caption{\textbf{Per-bias cosine matrices across all 5 instruct families.} Bottom-right block in each panel corresponds to the 4-bias cluster (\biasname{Distractor Fact}, \biasname{Spurious Few-Shot Squares}, \biasname{Wrong Few-Shot}, \biasname{Suggested Answer}). The cluster is dense on Qwen, moderate on Gemma, weak on Mistral, and not significant on Llama or OLMo, matching the rank/significance ordering in Table~\ref{tab:cluster-all}. The \biasname{Spurious Few-Shot Hindsight} singleton (``---'' entries on Gemma and OLMo) is not part of those models' per-bias cosine analysis.}
  \label{fig:cosine-side-by-side}
\end{figure*}

\begin{figure*}[!htbp]
  \centering
  \includegraphics[width=0.49\linewidth]{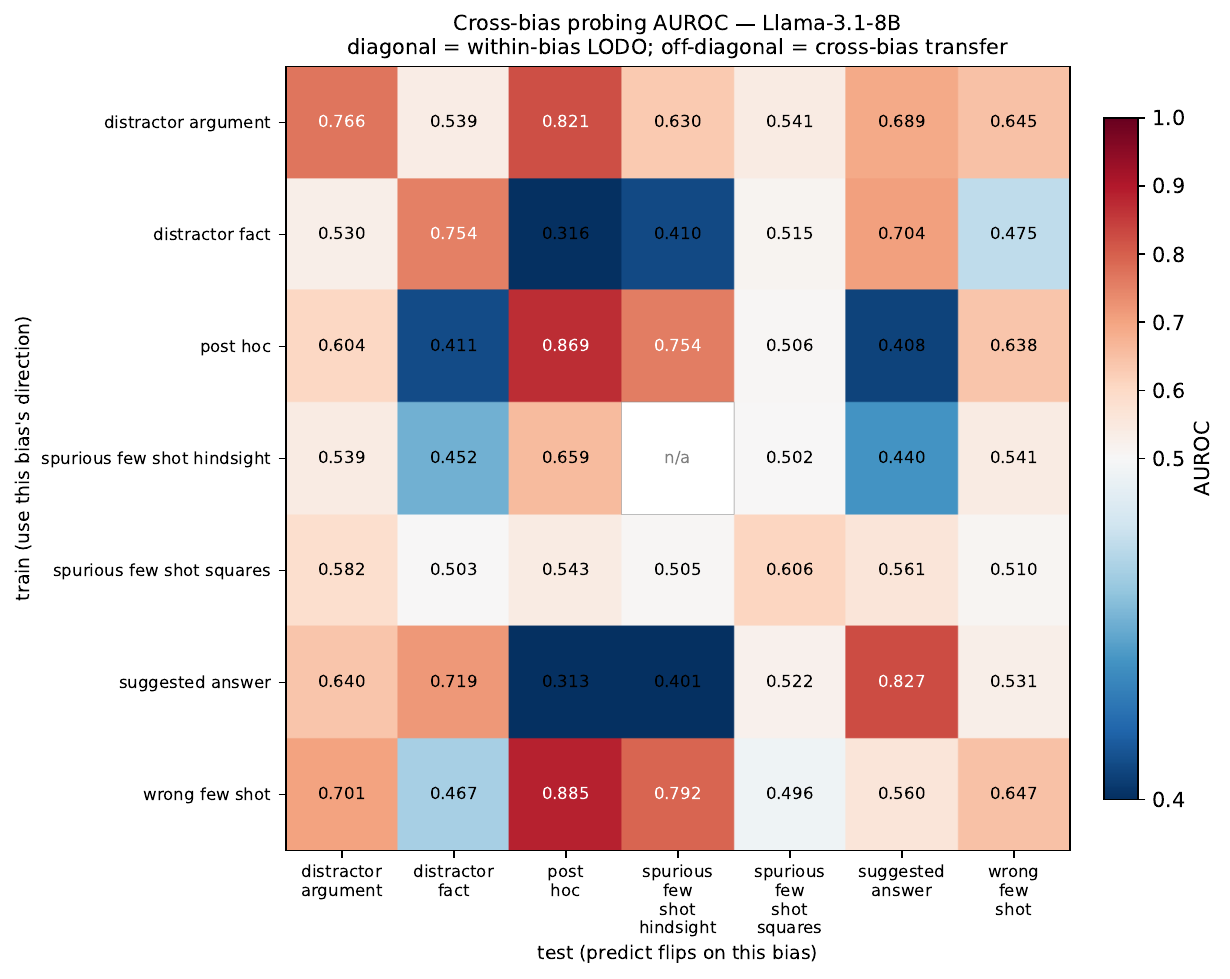}
  \hfill
  \includegraphics[width=0.49\linewidth]{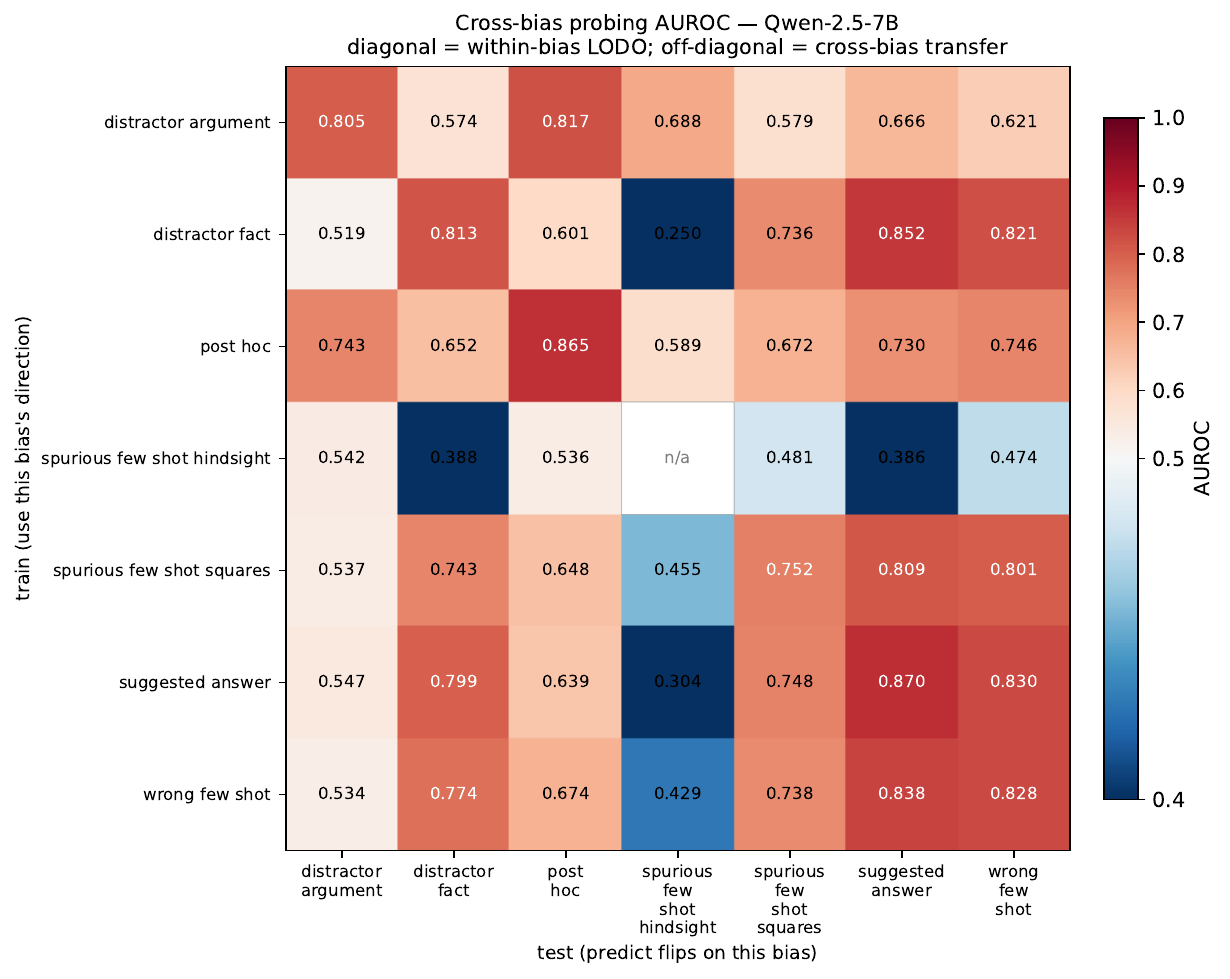}
  \caption{\textbf{Cross-bias probing transfer AUROC.} Rows are the bias whose direction is used as the probe (\emph{B\textsubscript{train}}); columns are the bias whose flipped/resisted pairs are predicted (\emph{B\textsubscript{test}}). Diagonal cells reproduce the within-bias LODO AUROC where available; the \biasname{Spurious Few-Shot Hindsight} diagonal is n/a because it has no within-bias LODO partner. Off-diagonal cells measure cross-bias direction transfer.}
  \label{fig:probe-transfer}
\end{figure*}

On Qwen, the per-bias cosine matrix shows a tight cluster of four biases (\biasname{Suggested Answer}, \biasname{Distractor Fact}, \biasname{Wrong Few-Shot}, \biasname{Spurious Few-Shot Squares}) whose pairwise cosines all fall in $[0.72, 0.85]$. Whether this strong overlap is universal turns out to matter: Table~\ref{tab:cluster-all} shows that it is not. The same four biases cluster strongly on Qwen ($p = 2.5\!\times\!10^{-38}$, rank 1 of 35 possible 4-subsets), weakly but significantly on Gemma ($p = 10^{-3}$, rank 3 of 15) and Mistral ($p = 0.02$, rank 6 of 35), and not significantly on Llama or OLMo. Whether biases share representation is therefore a property of the specific model rather than of the bias category. Figure~\ref{fig:cosine-side-by-side} shows the per-bias cosine matrices across all five families. Cross-bias probing shows the same contrast: mean within-cluster off-diagonal AUROC is 0.79 on Qwen and 0.55 on Llama (Figure~\ref{fig:probe-transfer}). To test whether one generic direction is sufficient, we also
train a pooled direction on the other biases using question-disjoint
training data. The held-out bias's own direction performs better
in every evaluated case on both Llama and Qwen, although pooled
transfer is stronger on Qwen.

\begin{table}[t]
  \centering
  \small
  \begin{tabular}{lcccc}
    \toprule
    \textbf{Family} & \textbf{Within} & \textbf{Outside} & \textbf{MW $p$} & \textbf{Rank} \\
    \midrule
    \textbf{Qwen}   & 0.63 & 0.17 & $\mathbf{2.5\!\times\!10^{-38}}$ & \textbf{1 / 35} \\
    Gemma           & 0.32 & 0.24 & $1.2\!\times\!10^{-3}$ & 3 / 15 \\
    Mistral         & 0.20 & 0.13 & 0.02 & 6 / 35 \\
    Llama           & 0.07 & 0.07 & 0.68 (n.s.) & 19 / 35 \\
    OLMo            & 0.25 & 0.29 & 0.98 (n.s.) & 8 / 15 \\
    \bottomrule
  \end{tabular}
  \caption{\textbf{The 4-bias cluster is strongest on Qwen, not universal.} Within vs.\ outside cosines for the candidate cluster. Rank is among all 4-bias subsets.}
  \label{tab:cluster-all}
\end{table}

A clean stress test on the distinctness side: \biasname{Post Hoc} (the model is shown its apparent prior answer) and \biasname{Suggested Answer} (the user inserts ``I think it's X'') are both \emph{answer-anchoring} behaviorally. Are they the same direction? In no family. The pairwise cosines ($-0.25$ Llama, $+0.34$ Qwen, $+0.40$ Gemma, $+0.12$ Mistral, $+0.33$ OLMo) sit at or below each model's own off-diagonal average in 4 of 5 families, and are anti-aligned on Llama. Different sources of anchoring can occupy different directions inside the network.

\paragraph{Cross-bias intervention transfer.} The geometric picture has a functional correlate: if we use bias $A$'s direction to debias bias $B$ ($A \neq B$), some of the recovery transfers. For this cross-bias experiment we use the within-bias recovery-maximizing strength per model ($\alpha=0.5$ on Llama, $\alpha=4$ on Qwen; differs from Table~\ref{tab:debias}'s preservation-constrained operating $\alpha$). On Llama, off-diagonal recovery averages 21.9\% versus 29.2\% diagonal (a target bias's own direction recovers about 1.3$\times$ better than a cross-bias direction), and 0.8\% for a matched random direction. On Qwen, off-diagonal recovery is 13.3\% versus 18.5\% diagonal (1.4$\times$ better) and 4.4\% random. So directions are not orthogonal in function either, consistent with the moderate off-diagonal cosines in the per-bias matrix. But the diagonal still wins on both models tested: each bias has its own most-effective direction.

The biases therefore do not collapse into one shared representation: cross-bias overlap is model-specific, and even behaviorally similar biases can occupy different directions.

\section{Direction as a Proof-of-Concept Debiasing Tool}
\label{sec:debias}

\S\ref{sec:category} used the intervention to show that the direction causally affects answer selection: subtracting it from the residual stream changes the model's answer. That was a yes/no test at a fixed strength. This section asks a different question: as a debiasing intervention, how much bias can we actually remove, and at what cost to correct answers? Same hook on the same direction, but a different success metric. We now sweep the steering strength $\alpha$ and pick a per-model operating point.

\paragraph{Operating point.} We sweep $\alpha \in \{0.1, 0.25, 0.5, 1, 2, 4, 8\}$ and, for each model, pick the $\alpha$ that maximizes mean recovery on flipped pairs subject to mean preservation on already-correct resisted pairs staying $\geq 90\%$. We use one global $\alpha$ per model.

\begin{table}[t]
  \centering
  \small
  \begin{tabular}{lcccc}
    \toprule
    \textbf{Model} & $\alpha^*$ & \textbf{Rec.} & \textbf{Pres.} & \textbf{vs rand} \\
    \midrule
    Qwen           & 4.00 & 20.1\% & 94.8\% & $4.3\times$ \\
    OLMo           & 0.25 & 15.7\% & 92.8\% & $9.9\times$ \\
    Gemma          & 8.00 & 14.6\% & 97.7\% & $13\times$ \\
    Mistral        & 0.10 & 13.2\% & 90.4\% & $8.9\times$ \\
    Llama          & 0.10 & 7.4\%  & 96.4\% & $31\times$ \\
    \modelname{Qwen-base}      & 1.00 & 5.5\%  & 93.6\% & $7.4\times$ \\
    \bottomrule
  \end{tabular}
  \caption{\textbf{Same intervention, modest debiasing effect with a transparent tradeoff.} For each model, $\alpha^*$ maximizes mean recovery (Rec.) subject to mean preservation (Pres.) $\geq 90\%$. ``vs rand'' = ratio of recovery with the data-derived direction
to recovery with a random direction at matched $\alpha$.}
  \label{tab:debias}
\end{table}

\paragraph{Selectivity on unbiased prompts.} A natural worry is that subtracting the bias direction damages general capability, not just the bias. Applying the intervention at operating $\alpha$ to the unbiased version of each held-out pair (same questions, bias cue stripped) keeps accuracy at 94--98\% across all 5 instruct families, versus 99\% under a random direction at matched $\alpha$: the intervention is direction-specific and clean accuracy stays high.

\paragraph{Per-bias picture.} Recovery is not uniform: content and anchoring biases (\biasname{Distractor Fact}, \biasname{Post Hoc}, \biasname{Suggested Answer}) recover at $\sim$15--20\%, format-pattern biases (\biasname{Wrong Few-Shot}, \biasname{Distractor Argument}) at $\sim$7--13\%. A natural follow-up question (whether bias $A$'s direction can debias bias $B$ where $A \neq B$) is covered in \S\ref{sec:distinct} (``Cross-bias intervention transfer''), since the answer (some transfer, but the diagonal always wins) speaks to representational distinctness rather than to the debiasing application.

The same intervention that establishes a causal effect also reduces the bias modestly: at a sensibly-chosen $\alpha$, it recovers 7--20\% of bias-induced errors while preserving $\geq 90\%$ of correct answers across all 5 instruct families. The intervention is bidirectional, though the two directions are asymmetric: subtracting along the bias direction (positive $\alpha$) lowers the bias rate by up to 42 percentage points, while pulling toward the bias mean (negative $\alpha$) raises it by a smaller margin (up to 7 points, Figure~\ref{fig:bidirectional}). The recovery-vs-preservation curves underlying the operating points are in Appendix~\ref{app:intervention-detail} (Figure~\ref{fig:intervention-tradeoff}).

\begin{figure}[t]
  \centering
  \includegraphics[width=\linewidth]{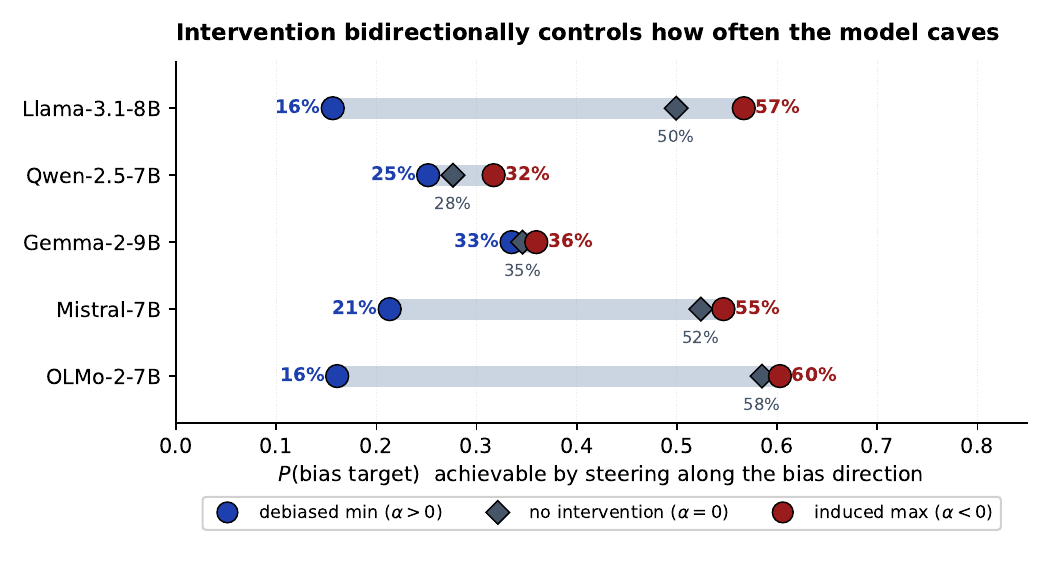}
  \caption{\textbf{Steering along the bias direction adds or removes bias.} Per model, the achievable range of $P(\mathrm{bias\ target})$ when sweeping intervention strength $\alpha \in [-2, 2]$: \textcolor[HTML]{1e40af}{debiased min} ($\alpha>0$, push away from the bias mean), \textcolor[HTML]{475569}{baseline} ($\alpha = 0$, no intervention), and \textcolor[HTML]{991b1b}{induced max} ($\alpha<0$, push toward the bias mean). The direction is causal in both directions, and the effect is largest on the models with highest baseline bias rates.}
  \label{fig:bidirectional}
\end{figure}

\section{Discussion}
\label{sec:discussion}

The literature largely treats cue-induced bias as a general failure mode of LLMs, and per-bias probing methods proceed as if each bias has its own dedicated machinery. Both deserve revision: the susceptibility is largely shaped by alignment tuning, and whether different biases share a direction is a property of the specific model, not of the bias category. Per-bias selectivity is an empirical model-specific property, not a guarantee of the construction. The base-probe (borrow instruct labels, decode from base activations) is a natural recipe for sanity-checking ``pre-existing'' vs.\ ``shaped by tuning'' for any behavioral phenomenon.

Representationally, the asymmetry in Table~\ref{tab:rebuttal} is suggestive: in base activations only the explicit-lexical-cue bias leaves the clearest trace, while the more abstract biases leave much weaker traces. One reading is that pretraining lets the model register an explicit cue, while alignment tuning\footnote{``Alignment tuning'' here covers whatever post-training (SFT, RLHF, DPO, or combinations) takes a base checkpoint to its instruct counterpart. For four families, we compare only the released base and instruct endpoints. For OLMo-2, we additionally analyze the released SFT and DPO checkpoints in \S\ref{sec:origin}.} amplifies and reorganizes how these cues influence answer selection. The pooled-direction test in \S\ref{sec:distinct} argues against one generic linear ``follow context'' direction being sufficient. Distinguishing broader shared mechanisms requires controlled fine-tuning experiments we leave to future work.

A practical implication is that cue-induced bias should not be treated merely as a prompt-engineering nuisance, or as a fixed weakness of the pretrained model. If the relevant directions are strongly amplified or reorganized during post-training, then mitigation should target alignment recipes themselves, not only inference-time defenses. Future alignment procedures could monitor whether instruction tuning increases sensitivity to irrelevant cues, spurious demonstrations, or fabricated conversational history. In this sense, cue-induced bias provides a useful diagnostic for post-training: a model can become more helpful and instruction-following while also becoming more prone to over-weighting the user's framing.

\section{Conclusion}
\label{sec:conclusion}

Across seven cue-induced biases and five model families, three converging measures (probing, LODO transfer, causal intervention) give one picture. Each bias has a coherent, decodable, causally effective linear direction. The representation is largely shaped by alignment tuning, and the biases do not collapse into a single shared representation, with cross-bias overlap varying across models. The same intervention also provides a POC debiasing tool. Cue-induced bias is therefore best understood as a family of causally effective linear directions that are largely shaped by alignment tuning.\footnote{Code and experiment artifacts: \url{https://github.com/prakharg55/bias-direction-EMNLP}}

\section*{Limitations}
\label{sec:limitations}

\paragraph{Non-CoT only.} We measure the hidden state at the moment of single-token answering. Many real-world bias manifestations involve longer reasoning chains. Our findings may not transfer to CoT scenarios where bias-following emerges during intermediate reasoning, and a CoT-pilot extension we ran (not included here) suggests the within-bias direction in the CoT setting is much weaker and largely orthogonal to the non-CoT one, a question for follow-up work. Because we study only the final prompt position, these experiments capture the answer-decision state rather than the full computation over earlier positions.

\paragraph{\biasname{Post Hoc} on OLMo is unfillable.} \modelname{OLMo-2-7B-Instruct} caves to \biasname{Post Hoc} at $\sim$99\% across datasets, leaving no LODO-testable resisted contrast in 3 of 4 datasets (two datasets have zero resisted pairs; a third has only 2, below our minimum-pair threshold). We report this as an honest behavioral finding (OLMo is unusually susceptible) rather than a missing experiment.

\paragraph{Excluded BCT biases.} \biasname{Positional Bias} is LLM-as-judge format, not MCQ. \biasname{Are You Sure} is a 5-turn dialogue in which a fake assistant turn writes a fully-articulated retraction paragraph, a chain of reasoning baked into the prompt that fundamentally violates our non-CoT framing.

\paragraph{Imperfect debiasing selectivity on some models.} Mistral and Llama have narrower preservation-recovery operating windows than Gemma and Qwen. The intervention is direction-specific everywhere (data-derived $\gg$ random at matched $\alpha$) but not uniformly clean. Conditional or soft steering (e.g., gated by a bias-detection signal) is a natural next step.

\paragraph{Model coverage.} We study five 7--9B model families, so broader coverage is needed to determine how general the cross-model patterns are.

\section*{Acknowledgments}
This material is based in part upon work supported
by the German Federal Ministry of Education and Research (BMBF): Tübingen AI Center, FKZ: 01IS18039B; by the Machine Learning Cluster of Excellence, EXC number 2064/1 – Project number 390727645; 
by Coefficient Giving; 
by Schmidt Sciences; 
by the Cooperative AI Foundation; by the Canadian AI Safety Institute Research Program at CIFAR. 
Resources used in preparing this research project were provided, in part, by the Digital Research Alliance of Canada; the Province of Ontario; the Swiss AI Compute grants; the Government of Canada through CIFAR; and companies sponsoring the Vector Institute.
F.D. acknowledges support through a fellowship from the Hector Fellow Academy.
We thank Jiarui Liu, Andrew Kim, and the other key developers of PaperMentor \citep{liu2026papermentorhumancenteredmultiagentwriting}, which provides mentorship and valuable suggestions in polishing this paper after the first draft by the authors. We further acknowledge the crucial work of AdminBot for administrative support that streamlined our research workflow.

\bibliography{custom}

\appendix

\section{Datasets and biases}
\label{app:datasets-biases}

\subsection{Bias type definitions}
\label{app:bias-defs}

We summarize the seven BCT biases used in this paper \citep{chua2024bias}. Each bias modifies the user message in a way that creates a (biased, unbiased) pair on the same underlying question. The model's susceptibility is measured by how often the biased version flips its answer.

\paragraph{\biasname{Suggested Answer}.} The user message ends with an explicit suggestion (``\emph{I think the answer is X}'') before the question, where X is the bias-target letter. This is the only bias in our set that is instruction-like in form: the cue appears in the user's voice as a direct suggestion.

\paragraph{\biasname{Distractor Argument}.} An irrelevant but plausible-sounding argument for a wrong answer is inserted into the user message before the question. The argument does not address the actual question content but provides surface-level rhetorical pressure toward the bias-target letter.

\paragraph{\biasname{Distractor Fact}.} An irrelevant but factually-true statement that points toward a wrong answer is inserted before the question. Unlike \biasname{Distractor Argument}, the inserted material is factually correct. What makes it biasing is its irrelevance to the actual question.

\paragraph{\biasname{Wrong Few-Shot}.} The prompt contains a few-shot block in which the demonstrations have systematically wrong labels (the demonstrated answers point toward the bias-target letter rather than the correct one).

\paragraph{\biasname{Spurious Few-Shot Squares}.} The few-shot demonstrations share a spurious surface feature (their question stems end with the word ``\emph{squares}'') that correlates with the bias-target letter in-context. The model is expected to pick up on the spurious shortcut.

\paragraph{\biasname{Spurious Few-Shot Hindsight}.} A singleton source built on the Hindsight Neglect dataset \citep{mckenzie2023inverse}: few-shot demonstrations follow a hindsight-bias pattern such that resisting the cue requires ignoring the in-context shortcut. Because it has only one underlying dataset, this bias has no within-bias LODO partner and is excluded from per-bias LODO tables.

\paragraph{\biasname{Post Hoc}.} A fake prior assistant turn is inserted into the dialogue, in which the assistant has apparently already committed to the bias-target letter. The user then asks the question, and the model is steered toward consistency with the fabricated ``previous'' answer. In our non-CoT framing this behaves as an answer-anchoring bias.

\paragraph{Excluded biases.} We exclude two BCT biases that are format-incompatible with our pipeline. \biasname{Positional Bias} is an LLM-as-judge format (not MCQ) and does not fit the single-token A--P decoding setup. \biasname{Are You Sure} is a five-turn dialogue in which a fake assistant turn writes a fully-articulated retraction paragraph. This embedded reasoning chain is incompatible with our non-CoT framing.

\subsection{Pair-count statistics}
\label{app:pair-stats}

After applying the filtering described in \S\ref{sec:method} (dropping pairs where the bias target coincides with the correct letter, or where the model did not know the unbiased answer), the per-model totals are in Table~\ref{tab:pair-counts-models} and the per-(bias, dataset) breakdowns are in Tables~\ref{tab:per-source-instruct} and \ref{tab:per-source-base}. The instruct totals match those used in Table~\ref{tab:base-flip} (rounded). The dramatic instruct-vs-base flip-rate disparity for four of five families, and the \modelname{Qwen-base} outlier, are discussed in \S\ref{sec:origin}.

\begin{table}[h]
  \centering
  \small
  \begin{tabular}{lrr}
    \toprule
    \textbf{Model} & \textbf{Flipped} & \textbf{Resisted} \\
    \midrule
    \modelname{Llama-3.1-8B-Instruct}      & 5{,}953 & 6{,}733 \\
    \modelname{Qwen2.5-7B-Instruct}        & 3{,}677 & 11{,}906 \\
    \modelname{Gemma-2-9B-it}              & 5{,}222 & 11{,}236 \\
    \modelname{Mistral-7B-Instruct-v0.3}   & 5{,}626 & 5{,}664 \\
    \modelname{OLMo-2-1124-7B-Instruct}    & 7{,}146 & 3{,}900 \\
    \midrule
    \modelname{Llama-3.1-8B} (base)        & 11    & 3{,}162 \\
    \modelname{Qwen2.5-7B} (base, outlier) & 5{,}580 & 4{,}668 \\
    \modelname{Gemma-2-9B} (base)          & 62    & 3{,}925 \\
    \modelname{Mistral-7B-v0.3} (base)     & 95    & 1{,}342 \\
    \modelname{OLMo-2-1124-7B} (base)      & 280   & 3{,}427 \\
    \bottomrule
  \end{tabular}
  \caption{\textbf{Aggregate strict-flipped and strict-resisted pair counts per model checkpoint, summed over all 25 (bias, dataset) sources.} Per-(bias, dataset) breakdowns are in Tables~\ref{tab:per-source-instruct} and~\ref{tab:per-source-base}.}
  \label{tab:pair-counts-models}
\end{table}

\begin{table*}[!t]
  \centering
  \scriptsize
  \setlength{\tabcolsep}{3pt}
  \begin{tabular}{llrrrrrrrrrr}
    \toprule
    \textbf{Bias} & \textbf{Dataset} & \multicolumn{2}{c}{\modelname{Llama}} & \multicolumn{2}{c}{\modelname{Qwen}} & \multicolumn{2}{c}{\modelname{Gemma}} & \multicolumn{2}{c}{\modelname{Mistral}} & \multicolumn{2}{c}{\modelname{OLMo}} \\
    \cmidrule(lr){3-4}\cmidrule(lr){5-6}\cmidrule(lr){7-8}\cmidrule(lr){9-10}\cmidrule(lr){11-12}
    & & F & R & F & R & F & R & F & R & F & R \\
    \midrule
    \biasname{Sugg.\ Answer} & HSwag & 478 & 657 & 159 & 1{,}050 & 208 & 1{,}055 & 405 & 556 & 418 & 467 \\
     & LogiQA & 92 & 117 & 47 & 200 & 58 & 174 & 89 & 88 & 103 & 112 \\
     & MMLU & 258 & 713 & 120 & 880 & 118 & 946 & 249 & 502 & 305 & 505 \\
     & TQA & 68 & 265 & 52 & 350 & 41 & 398 & 68 & 260 & 92 & 115 \\
    \addlinespace[2pt]
    \biasname{Post Hoc} & HSwag & 840 & 294 & 362 & 857 & 481 & 793 & 673 & 290 & 902 & 0 \\
     & LogiQA & 202 & 14 & 124 & 126 & 172 & 66 & 132 & 40 & 218 & 0 \\
     & MMLU & 689 & 294 & 247 & 746 & 391 & 684 & 480 & 253 & 807 & 26 \\
     & TQA & 218 & 124 & 64 & 352 & 122 & 327 & 171 & 146 & 212 & 2 \\
    \addlinespace[2pt]
    \biasname{Dist.\ Fact} & HSwag & 157 & 922 & 211 & 971 & 105 & 1{,}136 & 156 & 746 & 723 & 165 \\
     & LogiQA & 49 & 145 & 34 & 207 & 27 & 200 & 31 & 127 & 104 & 110 \\
     & MMLU & 120 & 775 & 104 & 875 & 65 & 974 & 76 & 617 & 360 & 430 \\
     & TQA & 42 & 264 & 54 & 340 & 21 & 395 & 23 & 277 & 128 & 65 \\
    \addlinespace[2pt]
    \biasname{Dist.\ Arg.} & HSwag & 411 & 176 & 535 & 113 & 406 & 258 & 441 & 67 & 437 & 38 \\
     & LogiQA & 110 & 33 & 153 & 27 & 120 & 41 & 117 & 13 & 125 & 21 \\
     & MMLU & 180 & 208 & 228 & 167 & 156 & 266 & 194 & 104 & 218 & 112 \\
     & TQA & 52 & 181 & 109 & 187 & 37 & 261 & 109 & 119 & 86 & 71 \\
    \addlinespace[2pt]
    \biasname{Wrong FS} & HSwag & 442 & 357 & 124 & 1{,}011 & 298 & 926 & 539 & 264 & 443 & 308 \\
     & LogiQA & 127 & 37 & 48 & 180 & 108 & 122 & 123 & 20 & 112 & 90 \\
     & MMLU & 388 & 418 & 110 & 841 & 260 & 786 & 385 & 278 & 465 & 320 \\
     & TQA & 134 & 145 & 38 & 344 & 68 & 367 & 174 & 137 & 127 & 56 \\
    \addlinespace[2pt]
    \biasname{Spur.\ Squares} & HSwag & 300 & 284 & 278 & 920 & 863 & 414 & 375 & 282 & 264 & 271 \\
     & LogiQA & 82 & 28 & 99 & 144 & 186 & 50 & 99 & 39 & 96 & 65 \\
     & MMLU & 379 & 189 & 285 & 697 & 645 & 420 & 306 & 324 & 303 & 395 \\
     & TQA & 102 & 66 & 84 & 307 & 266 & 177 & 136 & 110 & 98 & 68 \\
    \addlinespace[2pt]
    \biasname{Spur.\ Hind.} & Hind. & 33 & 27 & 8 & 14 & 0 & 0 & 75 & 5 & 0 & 88 \\
    \bottomrule
  \end{tabular}
  \caption{\textbf{Per-(bias, dataset) strict-flipped (F) and strict-resisted (R) pair counts on the five instruct checkpoints.} Bias-row totals over the four standard datasets (\biasname{Spurious Hindsight} is a singleton) sum to the instruct rows of Table~\ref{tab:pair-counts-models}. \modelname{OLMo}'s \biasname{Post Hoc} is the only (model, bias) pair with no resisted contrast on any standard dataset.}
  \label{tab:per-source-instruct}
\end{table*}

\begin{table*}[!t]
  \centering
  \scriptsize
  \setlength{\tabcolsep}{3pt}
  \begin{tabular}{llrrrrrrrrrr}
    \toprule
    \textbf{Bias} & \textbf{Dataset} & \multicolumn{2}{c}{\modelname{Llama-base}} & \multicolumn{2}{c}{\modelname{Qwen-base}} & \multicolumn{2}{c}{\modelname{Gemma-base}} & \multicolumn{2}{c}{\modelname{Mistral-base}} & \multicolumn{2}{c}{\modelname{OLMo-base}} \\
    \cmidrule(lr){3-4}\cmidrule(lr){5-6}\cmidrule(lr){7-8}\cmidrule(lr){9-10}\cmidrule(lr){11-12}
    & & F & R & F & R & F & R & F & R & F & R \\
    \midrule
    \biasname{Sugg.\ Answer} & HSwag & 2 & 159 & 135 & 460 & 2 & 289 & 6 & 52 & 0 & 282 \\
     & LogiQA & 0 & 52 & 48 & 78 & 3 & 72 & 3 & 33 & 12 & 60 \\
     & MMLU & 0 & 159 & 205 & 628 & 6 & 232 & 8 & 97 & 24 & 202 \\
     & TQA & 0 & 40 & 69 & 178 & 1 & 55 & 2 & 24 & 10 & 48 \\
    \addlinespace[2pt]
    \biasname{Post Hoc} & HSwag & 0 & 279 & 827 & 21 & 0 & 221 & 0 & 26 & 6 & 367 \\
     & LogiQA & 1 & 74 & 170 & 4 & 0 & 60 & 1 & 50 & 3 & 76 \\
     & MMLU & 1 & 235 & 843 & 69 & 3 & 287 & 1 & 119 & 38 & 257 \\
     & TQA & 3 & 44 & 251 & 24 & 0 & 84 & 5 & 13 & 24 & 50 \\
    \addlinespace[2pt]
    \biasname{Dist.\ Fact} & HSwag & 0 & 210 & 321 & 524 & 12 & 362 & 9 & 53 & 3 & 338 \\
     & LogiQA & 1 & 76 & 72 & 92 & 0 & 81 & 10 & 41 & 1 & 51 \\
     & MMLU & 1 & 165 & 209 & 666 & 5 & 274 & 16 & 113 & 28 & 188 \\
     & TQA & 0 & 22 & 69 & 192 & 0 & 96 & 10 & 24 & 14 & 24 \\
    \addlinespace[2pt]
    \biasname{Dist.\ Arg.} & HSwag & 0 & 33 & 423 & 34 & 20 & 30 & 0 & 9 & 0 & 0 \\
     & LogiQA & 0 & 3 & 126 & 3 & 5 & 5 & 0 & 11 & 0 & 0 \\
     & MMLU & 0 & 4 & 351 & 36 & 1 & 3 & 0 & 5 & 0 & 0 \\
     & TQA & 0 & 1 & 191 & 26 & 0 & 2 & 0 & 6 & 0 & 0 \\
    \addlinespace[2pt]
    \biasname{Wrong FS} & HSwag & 0 & 329 & 191 & 323 & 0 & 352 & 1 & 76 & 3 & 340 \\
     & LogiQA & 1 & 77 & 49 & 57 & 0 & 81 & 1 & 56 & 2 & 66 \\
     & MMLU & 1 & 274 & 247 & 242 & 2 & 285 & 3 & 140 & 29 & 217 \\
     & TQA & 0 & 95 & 74 & 68 & 0 & 91 & 4 & 33 & 17 & 26 \\
    \addlinespace[2pt]
    \biasname{Spur.\ Squares} & HSwag & 0 & 314 & 178 & 403 & 0 & 376 & 0 & 71 & 5 & 368 \\
     & LogiQA & 0 & 73 & 62 & 64 & 0 & 85 & 1 & 56 & 3 & 76 \\
     & MMLU & 0 & 273 & 322 & 361 & 2 & 299 & 7 & 122 & 33 & 265 \\
     & TQA & 0 & 90 & 87 & 97 & 0 & 122 & 7 & 31 & 25 & 45 \\
    \addlinespace[2pt]
    \biasname{Spur.\ Hind.} & Hind. & 0 & 81 & 60 & 18 & 0 & 81 & 0 & 81 & 0 & 81 \\
    \bottomrule
  \end{tabular}
  \caption{\textbf{Per-(bias, dataset) strict-flipped (F) and strict-resisted (R) pair counts on the five pretrained base checkpoints.} Sums match the base rows of Table~\ref{tab:pair-counts-models}. Four of five base families almost never flip; \modelname{Qwen-base} is the documented behavioral outlier (\S\ref{sec:origin}). The \modelname{Dist.\ Arg.} block is near-empty on every base family except Qwen because base models without an instruction template rarely produce a parseable answer letter when shown a misleading argument.}
  \label{tab:per-source-base}
\end{table*}

\section{Per-bias direction probing}
\label{app:probing}

\begin{figure}[tb]
  \centering
  \includegraphics[width=\linewidth]{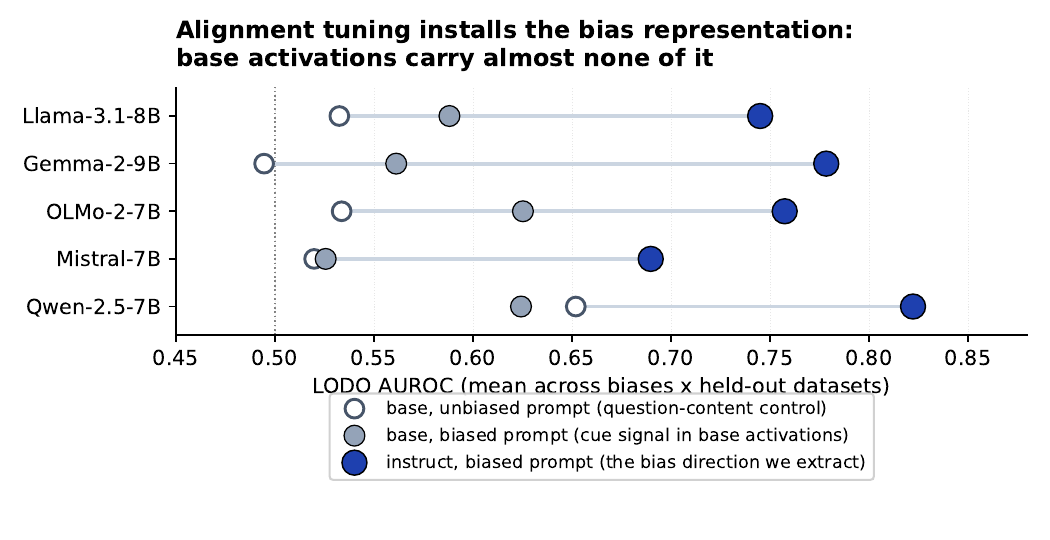}
  \caption{\textbf{The linearly accessible bias signal is much stronger after alignment tuning.} Per model, three LODO AUROC values on one axis: (white) the unbiased-prompt probe on base activations, a question-content control, (gray) the biased-prompt probe on base activations, asking whether the bias cue adds anything, and (blue) the same biased-prompt probe on instruct activations. White and gray sit near each other in every family, meaning the bias cue barely shifts base activations beyond question content. Blue is far to the right, meaning the bias direction is present in instruct activations. Numerical gaps are summarized in Table~\ref{tab:rebuttal} and the surrounding text.}
  \label{fig:cue-gap}
\end{figure}

\subsection{Layer-wise emergence}
\label{app:layer-wise}

\begin{figure*}[!htbp]
  \centering
  \includegraphics[width=0.49\linewidth]{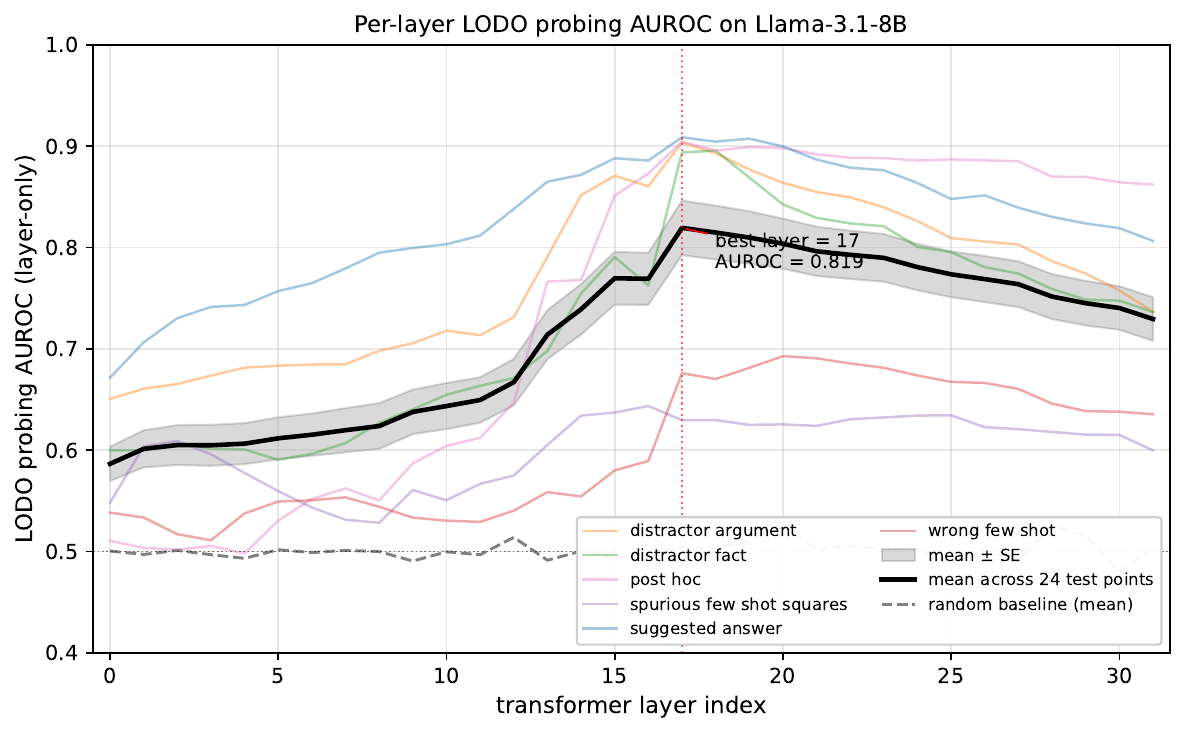}
  \hfill
  \includegraphics[width=0.49\linewidth]{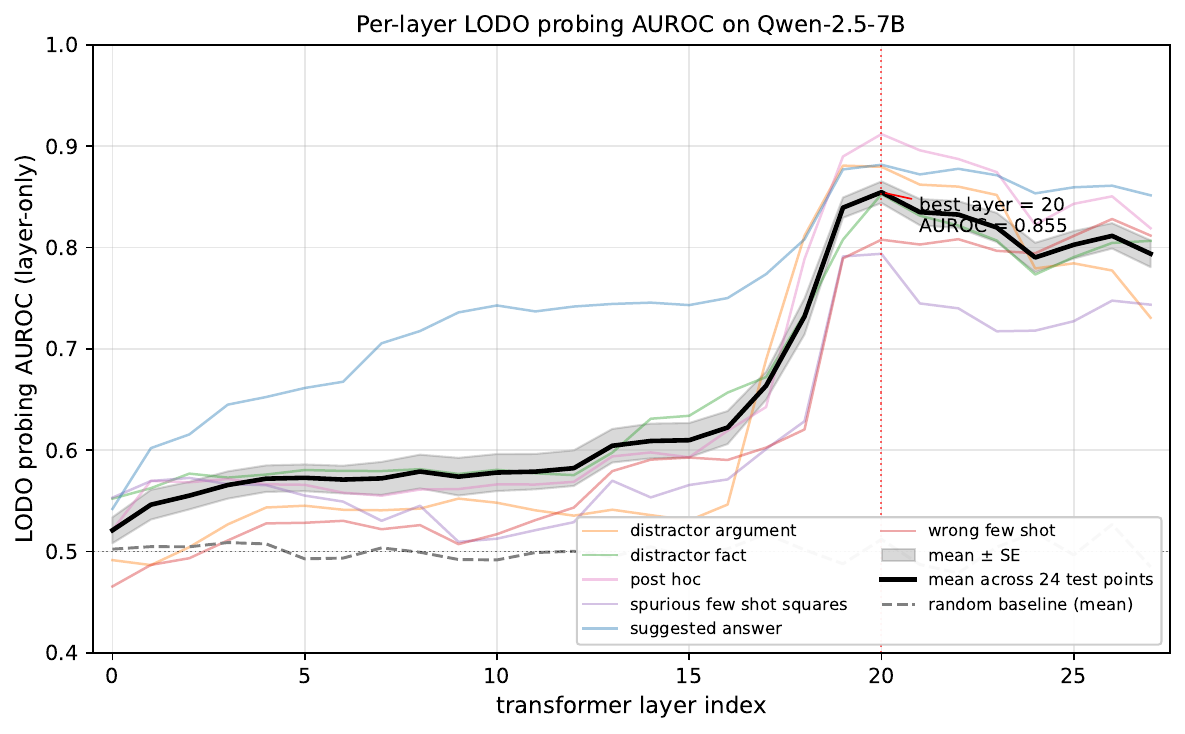}
  \caption{\textbf{Per-layer LODO probing AUROC on each transformer layer's hidden-state slice.} Solid black line is the mean across the 24 LODO test points (6 biases $\times$ 4 datasets). Gray band is $\pm 1$ standard error. Colored thin lines show per-bias means. Gray dashed line is the random-direction baseline. Red dotted vertical line marks the best layer.}
  \label{fig:layer-wise}
\end{figure*}

We slice the per-source bias direction into per-layer chunks $d_{s,\ell} \in \mathbb{R}^{H}$, where $H$ is the hidden size, and repeat the LODO probing setup at each layer $\ell$ in isolation: for each held-out source, the LODO direction is the unit-normalized mean of the per-source layer-$\ell$ directions across the other datasets of the same bias. Slicing a normalized vector and renormalizing the slice is mathematically equivalent to computing the layer-$\ell$ direction from $X_{\text{biased}}$ directly (both equal $(\mu^{\text{flip}}_\ell - \mu^{\text{rsst}}_\ell) / \|\mu^{\text{flip}}_\ell - \mu^{\text{rsst}}_\ell\|$, where $\mu_\ell$ is the layer-$\ell$ slice of the class-conditional mean). The per-layer random-direction baseline uses 20 random unit vectors in $\mathbb{R}^{H}$ (not in $\mathbb{R}^{LH}$, which would have a different chance level).

Figure~\ref{fig:layer-wise} plots the result for \modelname{Llama-3.1-8B} and \modelname{Qwen-2.5-7B}; the cross-family summary is in Figure~\ref{fig:layer-emergence}. On both models the AUROC rises through the network and peaks in the late-middle layers (Llama at layer $17/32$, 55\% of depth, AUROC $0.82$; Qwen at layer $20/28$, 74\% of depth, AUROC $0.85$). The best single-layer AUROC exceeds the all-layer LODO AUROC (Llama: $0.82$ vs $0.74$; Qwen: $0.85$ vs $0.82$), consistent with early layers adding noise to the all-layer concatenated direction.

\subsection{Letter-balanced control}
\label{app:letter-balanced}

A natural concern is that the bias direction might encode ``predict letter $X$'' rather than ``follow the bias,'' because the BCT data has non-uniform letter distributions for \texttt{bias\_target} (in flipped pairs) and \texttt{correct\_letter} (in resisted pairs). If so, the direction would partially capture a letter-prediction signal correlated with bias-following by construction.

We test this directly. For each source, we strictly balance the flipped subset by \texttt{bias\_target} letter (subsampling each letter group to the smallest group's size) and the resisted subset by \texttt{correct\_letter}, then recompute the bias direction from these balanced subsets and rerun the LODO probing test with both the averaged training direction \emph{and} the held-out test pairs taken from the balanced subsets. Strict balancing requires at least 3 flipped and 3 resisted pairs per letter; the number of LODO test points satisfying this varies by model (Table~\ref{tab:letter-balanced}). The unbalanceable sources are mostly \texttt{truthfulqa}, where letter distributions are highly imbalanced.

\begin{table*}[h]
  \centering
  \small
  \begin{tabular}{lccccc}
    \toprule
                            & \textbf{Llama} & \textbf{Qwen} & \textbf{Gemma} & \textbf{Mistral} & \textbf{OLMo} \\
    \midrule
    $n$ test pts            & 11             & 18            & 18             & 16               & 13             \\
    Orig AUROC              & 0.751          & 0.828         & 0.777          & 0.694            & 0.747          \\
    Balanced AUROC          & 0.726          & 0.820         & 0.818          & 0.661            & 0.721          \\
    Mean $\Delta$           & $-0.025$       & $-0.008$      & $+0.041$       & $-0.032$         & $-0.026$       \\
    Mean cosine             & $+0.908$       & $+0.908$      & $+0.843$       & $+0.725$         & $+0.721$       \\
    Wilcoxon $p$            & 0.15 (n.s.)    & 0.44 (n.s.)   & 0.11 (n.s.)    & 0.23 (n.s.)      & 0.13 (n.s.)    \\
    \bottomrule
  \end{tabular}
  \caption{\textbf{Letter-balanced control across all 5 instruct families.} After strictly balancing both the \texttt{bias\_target} distribution in flipped pairs and the \texttt{correct\_letter} distribution in resisted pairs, LODO AUROC changes by $|\Delta| \leq 0.041$ on average and the change is not statistically significant on any model. The balanced direction has cosine similarity $\geq 0.72$ with the original direction. The bias direction is not primarily a letter-prediction artifact in any family.}
  \label{tab:letter-balanced}
\end{table*}

On all 5 instruct families, the balanced AUROC differs from the original by $|\Delta| \leq 0.041$ on average, and the change is not statistically significant in any family ($p > 0.10$ throughout). Gemma is the lone case where balancing very slightly improves mean AUROC ($\Delta = +0.041$, still n.s.\ at $p = 0.11$); the other four families have small negative $\Delta$ values, consistent with a small but non-significant letter-prediction component in the unbalanced direction. The balanced direction is highly similar to the original everywhere (cosine $0.72$ on the weakest model, $0.91$ on the strongest). These results support that the direction primarily tracks bias-following rather than answer-letter identity across all five families.

\section{Cluster structure analysis}
\label{app:cluster-analysis}

\subsection{Per-source and per-bias cosine matrices}
\label{app:cosine-matrices}

\begin{figure*}[!htbp]
  \centering
  \includegraphics[width=0.49\linewidth]{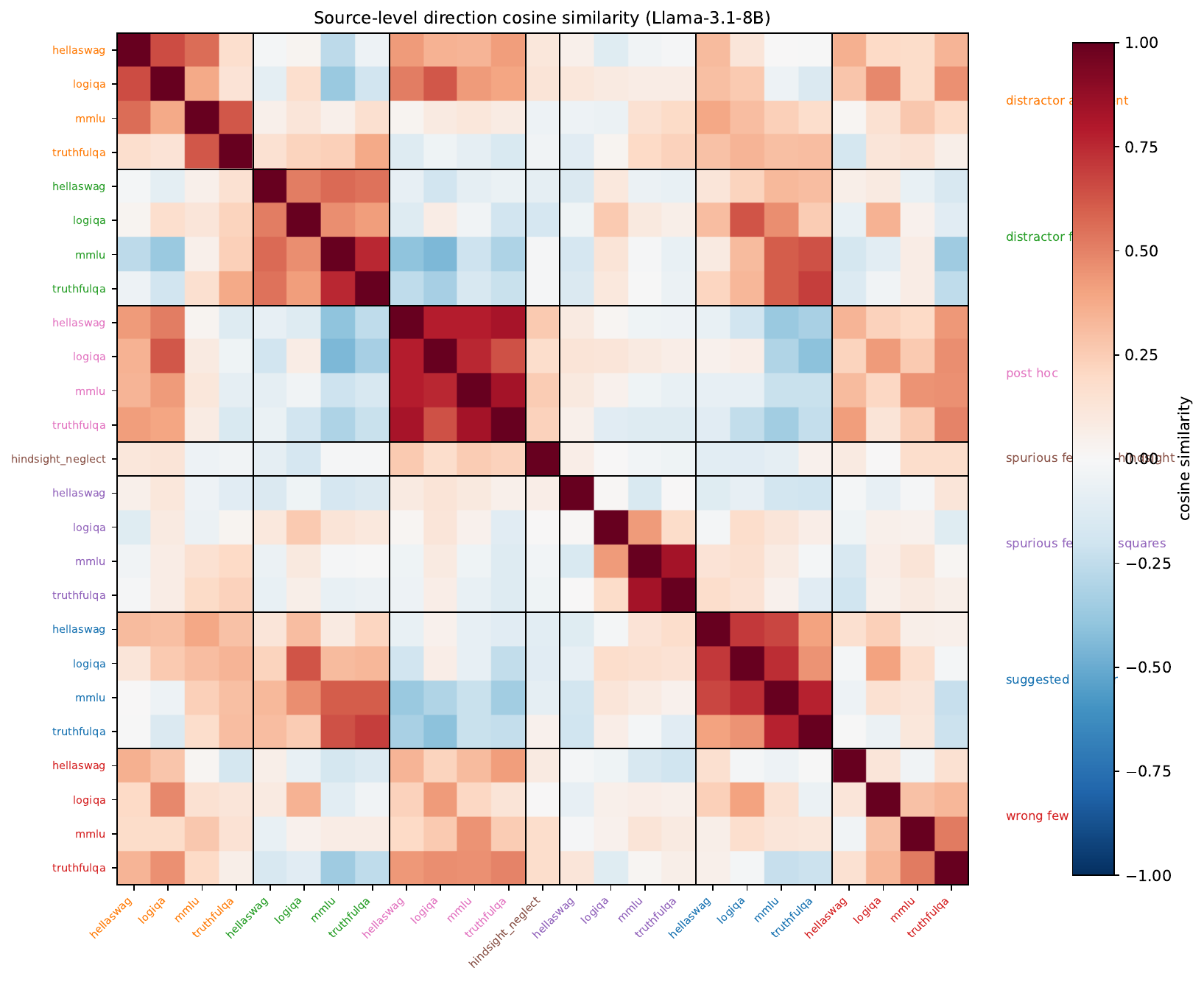}
  \hfill
  \includegraphics[width=0.49\linewidth]{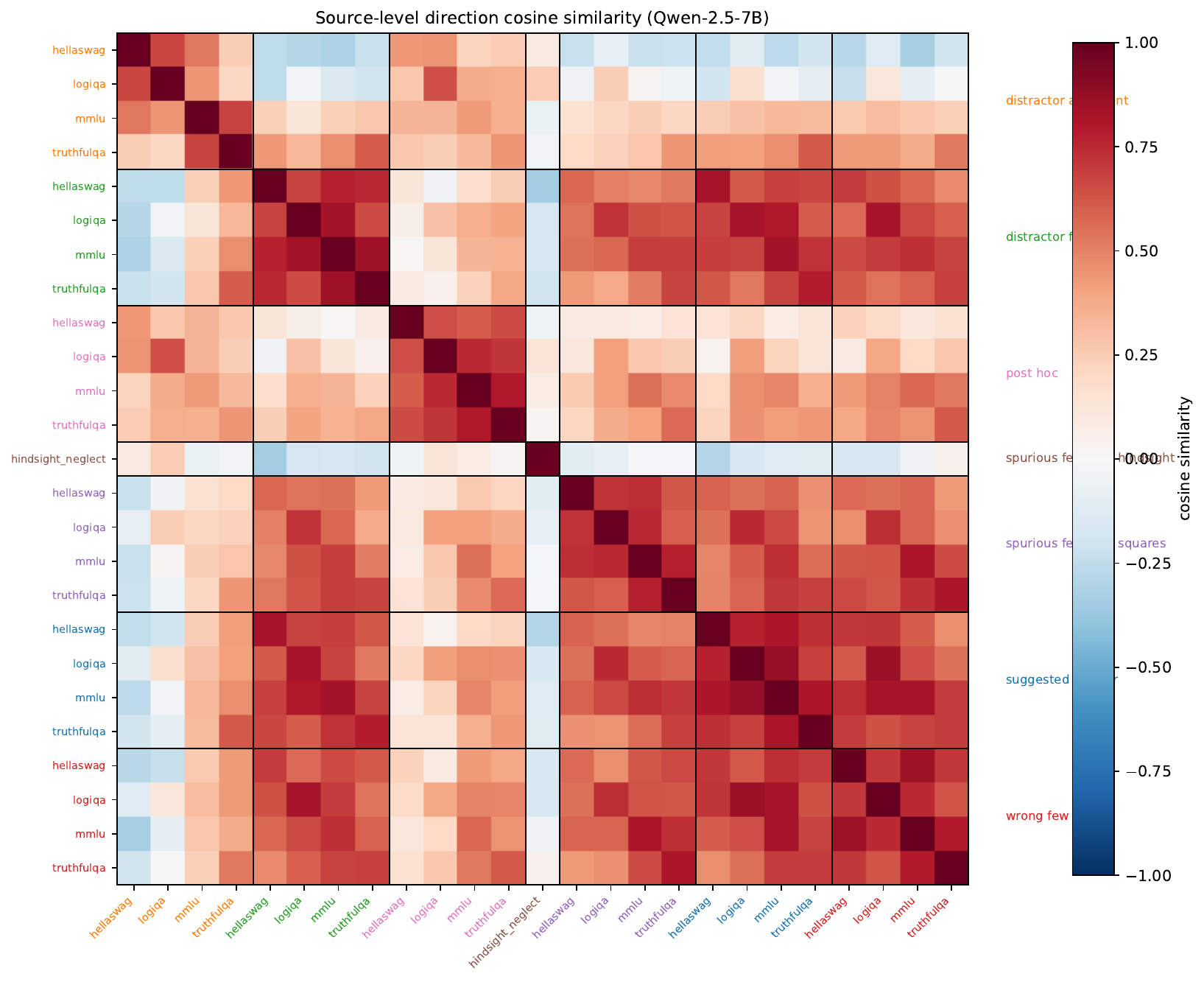}
  \caption{\textbf{Pairwise cosine similarity between the 25 per-source bias directions on \modelname{Llama-3.1-8B} (left) and \modelname{Qwen-2.5-7B} (right), reordered by bias type with bias-block boundaries marked.} Tick labels are colored by bias type. Within-bias blocks along the diagonal are visibly more correlated than off-block cells on both models.}
  \label{fig:source-cosine}
\end{figure*}

The five-family per-bias matrix is in Figure~\ref{fig:cosine-side-by-side}; the full $25 \times 25$ per-source matrix on Llama and Qwen is in Figure~\ref{fig:source-cosine}. With 6 non-singleton biases at 4 datasets each, the source matrix has $6 \cdot \binom{4}{2} = 36$ within-bias pairs and $\binom{25}{2} - 36 = 264$ across-bias pairs.

\paragraph{Within- vs across-bias separation.} Within-bias source-direction means are well above across-bias means in both models: Llama $+0.467$ vs $+0.068$ (diff $+0.399$, Mann-Whitney $U$ one-sided $p = 1.3 \times 10^{-12}$), Qwen $+0.689$ vs $+0.337$ (diff $+0.351$, $p = 6.2 \times 10^{-13}$). The within-vs-across lift is similar across models, but Qwen's absolute cosines are uniformly higher, consistent with the LODO finding that Qwen's bias representations are more concentrated. Per-bias breakdowns are in Table~\ref{tab:within-cosines}.

\paragraph{Cluster significance.} We sub-partition the 264 across-bias pairs to test whether Qwen's four-bias cluster is statistically distinguishable. A pair is \emph{within-cluster} if both sources' biases lie in the cluster (\biasname{Suggested Answer}, \biasname{Distractor Fact}, \biasname{Wrong Few-Shot}, \biasname{Spurious Few-Shot Squares}, 96 pairs), and the remaining 168 across-bias pairs are \emph{outside}. On Qwen the within-cluster mean cosine is $+0.634$ vs.\ $+0.168$ outside ($p = 2.5 \times 10^{-38}$); on Llama the same four-bias set shows no separation ($+0.068$ vs.\ $+0.068$, $p = 0.68$). The Llama result is a genuine negative control: the cluster is defined from Qwen's matrix, not Llama's, so the four-bias set is arbitrary for Llama. Among all $\binom{7}{4}=35$ four-bias subsets, Qwen's cluster has the highest mean within-subset across-bias cosine (rank 1/35), while the same set ranks 19/35 on Llama (Table~\ref{tab:cluster-all}). Because the cluster was identified from this same cosine matrix, the Qwen $p$-value is descriptive rather than confirmatory; the cluster's independent confirmation comes from the probing-transfer matrix (Appendix~\ref{app:cross-bias-probing}).

\paragraph{Per-bias and per-pair breakdowns.} Table~\ref{tab:within-cosines} gives per-bias within-source means, and Table~\ref{tab:top-pairs} lists the six pairs whose tight Qwen cluster (cosines $0.72$--$0.85$) is absent on Llama (only $1/6$ exceeds $0.10$).

\begin{table}[h]
  \centering
  \small
  \begin{tabular}{lcc}
    \toprule
    \textbf{Bias} & \textbf{Llama} & \textbf{Qwen} \\
    \midrule
    \biasname{Suggested Answer}            & +0.622 & +0.781 \\
    \biasname{Post Hoc}                    & +0.771 & +0.693 \\
    \biasname{Distractor Fact}             & +0.546 & +0.755 \\
    \biasname{Distractor Argument}         & +0.419 & +0.461 \\
    \biasname{Wrong Few-Shot}              & +0.230 & +0.739 \\
    \biasname{Spurious Few-Shot Squares}   & +0.217 & +0.702 \\
    \bottomrule
  \end{tabular}
  \caption{Per-bias mean within-source cosine similarity (6 pairs per bias).}
  \label{tab:within-cosines}
\end{table}

\begin{table}[h]
  \centering
  \footnotesize
  \setlength{\tabcolsep}{4pt}
  \begin{tabular}{@{}lcc@{}}
    \toprule
    \textbf{Pair} & \textbf{Llama} & \textbf{Qwen} \\
    \midrule
    \biasname{Dist Fact} $\leftrightarrow$ \biasname{Sugg Answer}       & +0.556 & +0.850 \\
    \biasname{Sugg Answer} $\leftrightarrow$ \biasname{Wrong Few-Shot}  & +0.099 & +0.836 \\
    \biasname{Dist Fact} $\leftrightarrow$ \biasname{Wrong Few-Shot}    & -0.095 & +0.789 \\
    \biasname{Spur Squares} $\leftrightarrow$ \biasname{Sugg Answer}    & +0.042 & +0.733 \\
    \biasname{Spur Squares} $\leftrightarrow$ \biasname{Wrong Few-Shot} & -0.002 & +0.783 \\
    \biasname{Dist Fact} $\leftrightarrow$ \biasname{Spur Squares}      & -0.003 & +0.716 \\
    \bottomrule
  \end{tabular}
  \caption{Cosine similarity for the six pairs that form Qwen's 4-bias cluster. On Qwen all six are in the 0.72--0.85 range; on Llama only one exceeds 0.10.}
  \label{tab:top-pairs}
\end{table}

\subsection{Cross-bias probing transfer}
\label{app:cross-bias-probing}

Figure~\ref{fig:probe-transfer} in the main text shows the full
$7\times7$ cross-bias probing matrix on Llama and Qwen.
\biasname{Spurious Few-Shot Hindsight} is included for cross-bias
transfer but has no within-bias LODO diagonal, which is shown as
n/a. The available diagonals reproduce the within-bias LODO AUROC from Section~\ref{sec:category} to fp32 precision. On Qwen the four cluster biases' directions are nearly interchangeable as probes (within-cluster off-diagonal AUROCs in $[0.74, 0.85]$, mean $0.79$). On Llama the same cluster does not transfer (within-cluster off-diagonal range $[0.47, 0.72]$, mean $0.55$).

Each cell carries an SE-based 95\% CI over its held-out test sources (mean $\pm 1.96 \cdot \text{SE}$ across the $n$ held-out sources of $B_{\text{test}}$). On Qwen all 12 within-cluster off-diagonal cells have 95\% CI lower bounds strictly above the 0.5 chance level, with CI half-widths $\leq 0.05$. On Llama only 2 of 12 do. Llama's above-chance cells that do exist are scattered across non-cluster pairs (7 of 18) rather than concentrated in any block, consistent with the absence of a cluster.

\section{Causal intervention details}
\label{app:intervention}

\subsection{$\alpha$-sweep}
\label{app:intervention-detail}

\begin{figure*}[!htbp]
  \centering
  \includegraphics[width=0.49\linewidth]{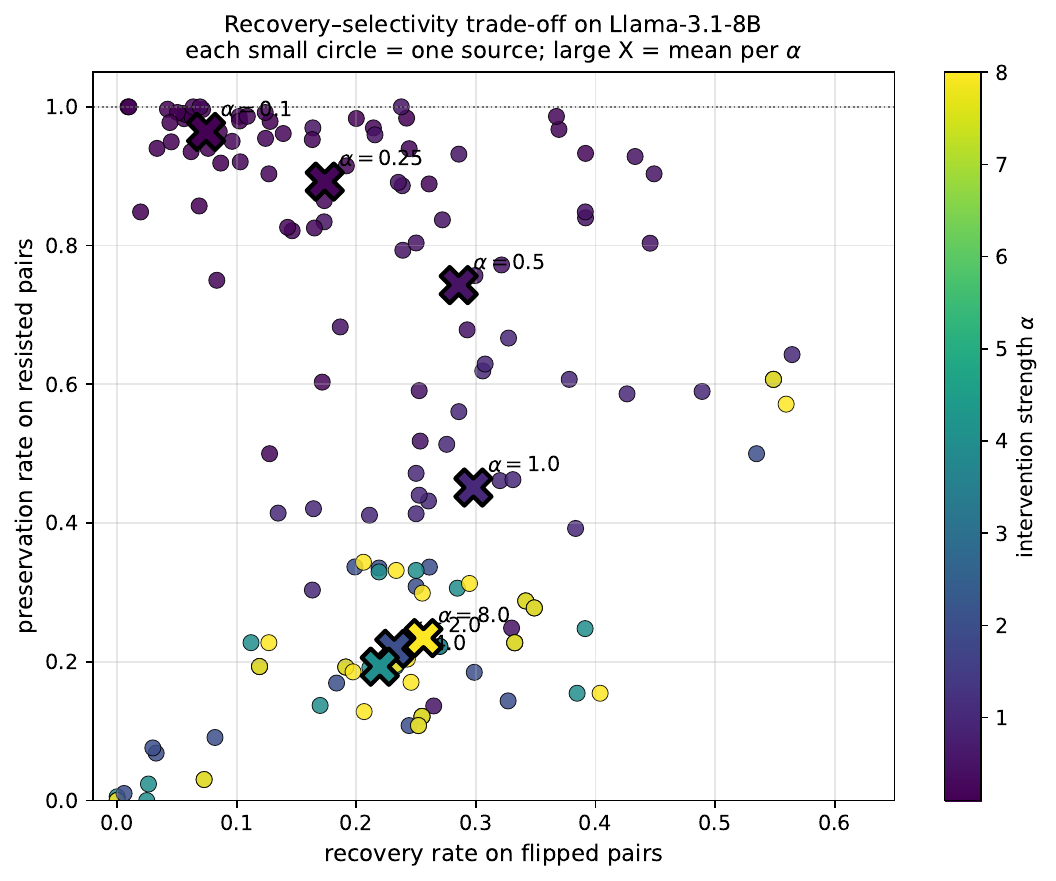}
  \hfill
  \includegraphics[width=0.49\linewidth]{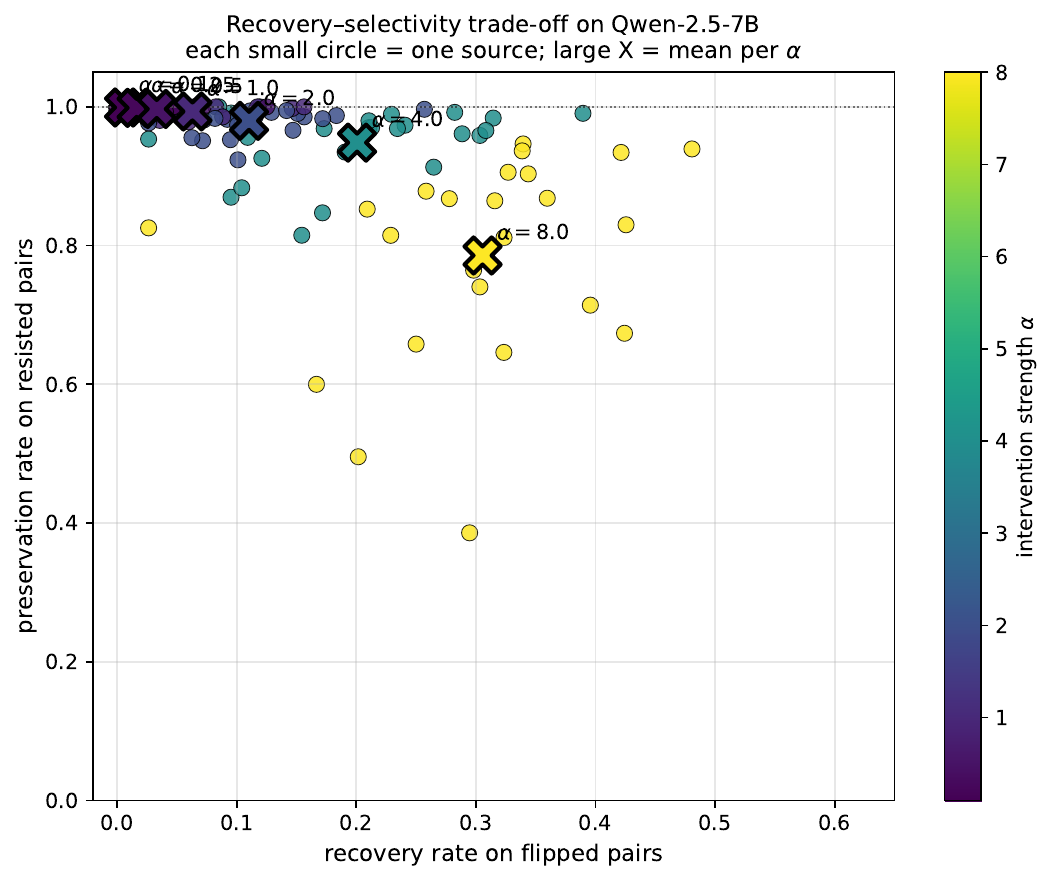}
  \caption{\textbf{Recovery--selectivity trade-off on \modelname{Llama-3.1-8B} (left) and \modelname{Qwen-2.5-7B} (right).} Each small dot is one (source, $\alpha$) combination for positive $\alpha$; large $\times$ markers are means per $\alpha$. On Qwen the curve is monotone, and preservation degrades only slightly through $\alpha = 4$.}
  \label{fig:intervention-tradeoff}
\end{figure*}

The intervention protocol is defined in Section~\ref{sec:debias}.
The operating point $\alpha^*$ in Table~\ref{tab:debias} is selected
separately for each model by maximizing mean recovery subject to
mean preservation $\geq 90\%$. Figure~\ref{fig:intervention-tradeoff}
shows the recovery--preservation sweep for
\modelname{Llama-3.1-8B} and \modelname{Qwen-2.5-7B} across
$\alpha \in \{0.1, 0.25, 0.5, 1, 2, 4, 8\}$; the bidirectional
summary on all five families is in Figure~\ref{fig:bidirectional}
(main body).

\subsection{Per-source recovery}
\label{app:per-source-intervention}

For this per-source diagnostic, we use $\alpha=0.5$ on
\modelname{Llama-3.1-8B}. This differs from the
preservation-constrained operating point $\alpha^*=0.1$
used for Llama in Table~\ref{tab:debias}.

\begin{figure*}[!htbp]
  \centering
  \includegraphics[width=0.95\linewidth]{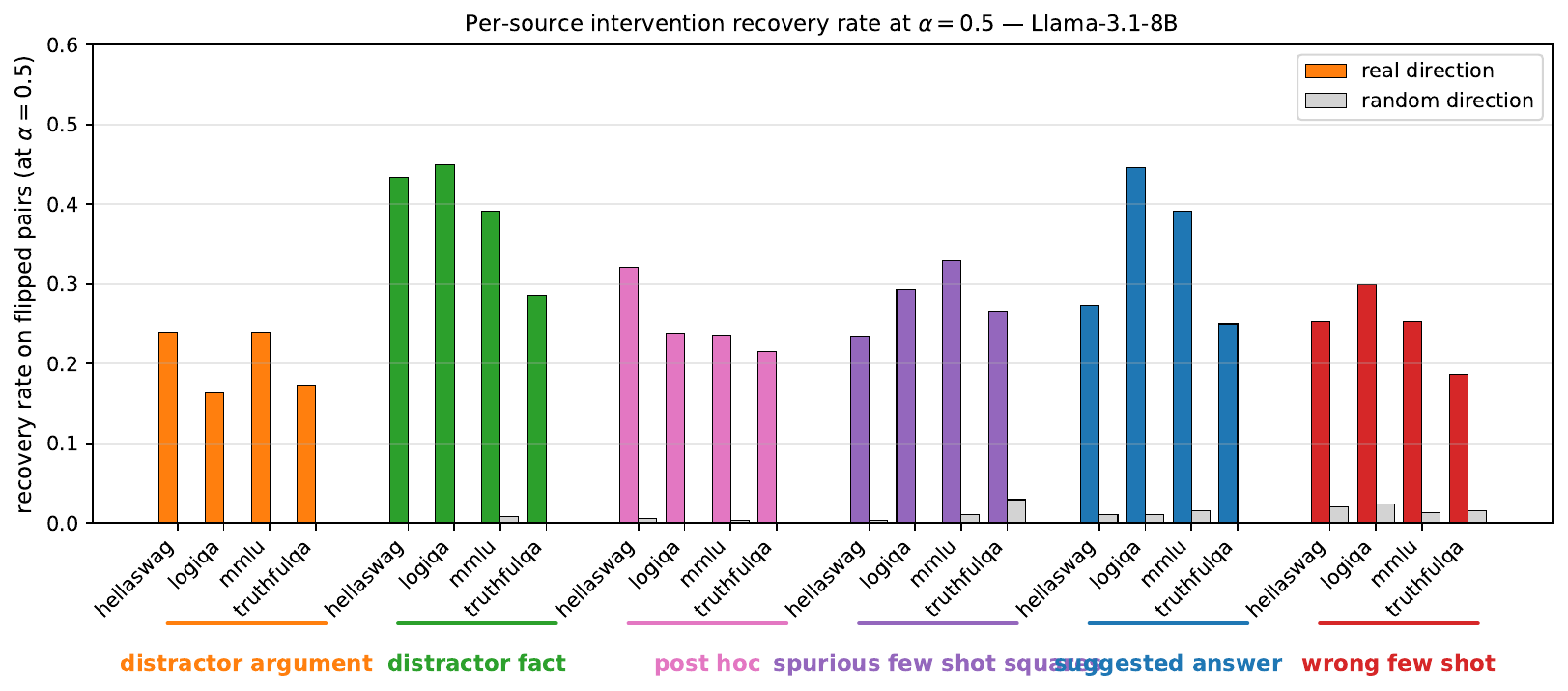}
  \caption{\textbf{Per-source intervention recovery at $\alpha = 0.5$ on \modelname{Llama-3.1-8B}.} All 24 sources show data-derived-direction recovery substantially exceeding random-direction recovery. \biasname{Distractor Fact} and \biasname{Suggested Answer} sources reach $40$--$45\%$ recovery. The random direction is essentially zero everywhere.}
  \label{fig:intervention-per-source}
\end{figure*}

Figure~\ref{fig:intervention-per-source} shows the per-source breakdown at $\alpha = 0.5$ for \modelname{Llama-3.1-8B}. All 24 sources are above the random-direction baseline. Recovery rates range from $16\%$ (\texttt{distractor\_argument\_\_logiqa}) to $45\%$ (\texttt{distractor\_fact\_\_logiqa}); random-direction recovery never exceeds $3\%$ on any source.

\end{document}